%% file: 0395.tex
\documentclass[runningheads]{llncs}
\usepackage[width=122mm,left=12mm,paperwidth=146mm,height=193mm,top=12mm,paperheight=217mm]{geometry}

\usepackage{graphicx}
\usepackage{amsmath,amssymb} 
\usepackage{color}

\input includes

\begin{document}
\title{Repeatability Is Not Enough:\\ Learning Affine Regions via Discriminability} 

\titlerunning{Repeatability Is Not Enough: Learning Affine Regions via Discriminability}

\author{Dmytro Mishkin\orcidID{0000-0001-8205-6718} \and
Filip Radenovi\'c\orcidID{0000-0002-7122-2765} \and
Ji\v{r}i Matas\orcidID{0000-0003-0863-4844}
}
%
\authorrunning{D. Mishkin, F. Radenovi\'c and J. Matas}
%

\institute{Visual Recognition Group, Center for Machine Perception, FEE, CTU in Prague\\
\email{\{mishkdmy, filip.radenovic, matas\}@cmp.felk.cvut.cz}}
\maketitle              
%

\begin{abstract}
A method for learning local affine-covariant regions is presented.
We show that maximizing geometric repeatability does not lead to local regions, a.k.a features,
that are reliably matched and this necessitates descriptor-based learning.
We explore factors that influence such learning and registration: the loss function, descriptor type, geometric parametrization and the trade-off between matchability and geometric accuracy and propose 
 a novel hard negative-constant loss function for learning of affine regions.
 The affine shape estimator -- AffNet -- trained with the hard negative-constant loss outperforms the state-of-the-art in bag-of-words image retrieval and wide baseline stereo.
 The proposed training process does not require precisely geometrically aligned patches.
 The source codes and trained weights are available at \url{https://github.com/ducha-aiki/affnet}
\end{abstract}
\input intro
\section{Learning affine shape and orientation}
\label{sec:main}
\subsection{Affine shape parametrization}
A local affine frame is defined by 6 parameters of the affine matrix. Two form a translation vector $(x,y)$ which is given by the keypoint detector and in the rest of the paper we omit it and focus on the \emph{affine transformation} matrix $A$,
\begin{equation}
\label{eq:A}
A =\left(
\begin{array}
{@{\hspace{0pt}}c@{\hspace{2pt}}c@{\hspace{0pt}}}
a_{11} & a_{12}\\
a_{21} & a_{22}\\ 
\end{array}\right).\end{equation}
Among many possible decompositions of matrix $A$, we use the following  
\begin{equation}
\label{eq:A'}
A = \lambda R(\alpha) A'  =  \det{A} 
\left(
\begin{array}{@{\hspace{0pt}}c@{\hspace{6pt}}c@{\hspace{0pt}}}
\cos{\alpha} & \sin{\alpha} \\
-\sin{\alpha} & \cos{\alpha}\\
\end{array}\right) \left(
\begin{array}
{@{\hspace{0pt}}c@{\hspace{2pt}}c@{\hspace{0pt}}}
a'_{11} & 0\\
a'_{21} & a'_{22}\\ 
\end{array}\right),
\end{equation}
where $\lambda = \det{A}$ is the scale, $R(\alpha)$ the \emph{orientation} matrix and $A'$~\footnote{$A'$ has a (0,1) eigenvector, preserving the vertical direction.} is the
\emph{affine shape} matrix with $\det{A'}$ = 1.
$A'$ is decomposed into identity matrix $I$ and \emph{residual shape}~$A''$:
\begin{equation}
\label{eq:A''}
A' = I + A'' = 
 \left(
\begin{array}{@{\hspace{0pt}}c@{\hspace{2pt}}c@{\hspace{0pt}}}
a'_{11} & 0 \\
a'_{21} & a'_{22}\\
\end{array}\right) =
 \left(\begin{array}{@{\hspace{0pt}}c@{\hspace{6pt}}c@{\hspace{0pt}}}
1 & 0\\
0 & 1\\ 
\end{array}\right)  + 
\left(
\begin{array}{@{\hspace{0pt}}c@{\hspace{2pt}}c@{\hspace{0pt}}}
a''_{11} & 0\\
a''_{21} & a''_{22}\\
\end{array}\right) 
\end{equation}
We show that the different parameterizations of the affine transformation significantly influence the performance of CNN-based estimators of local geometry, see Table~\ref{tab:aff-shape-parametrization}.
 \begin{figure}[htb]
 \centering
 Initial positions\\
 \includegraphics[width=0.30\linewidth]{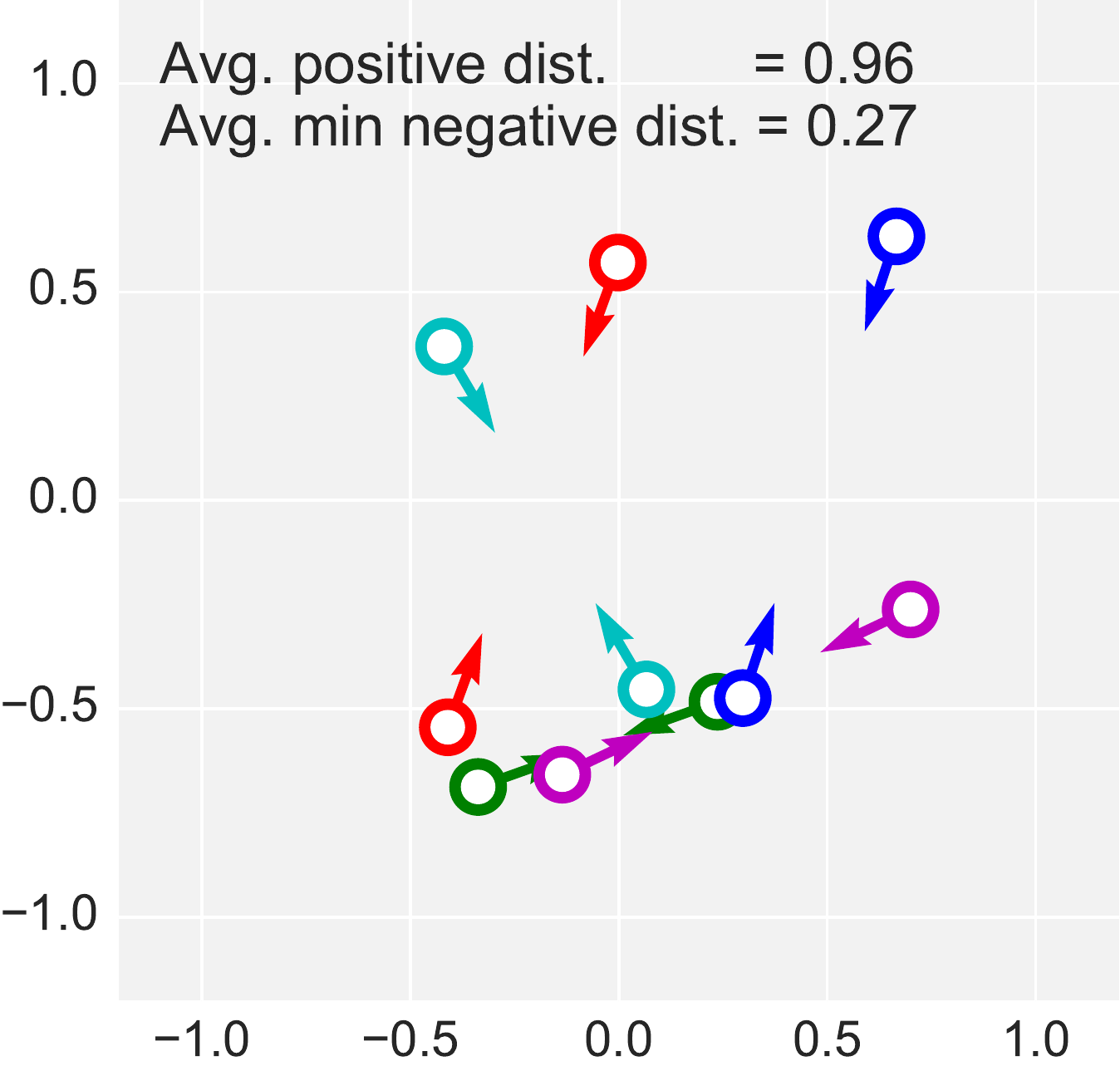}\hfill
 \includegraphics[width=0.30\linewidth]{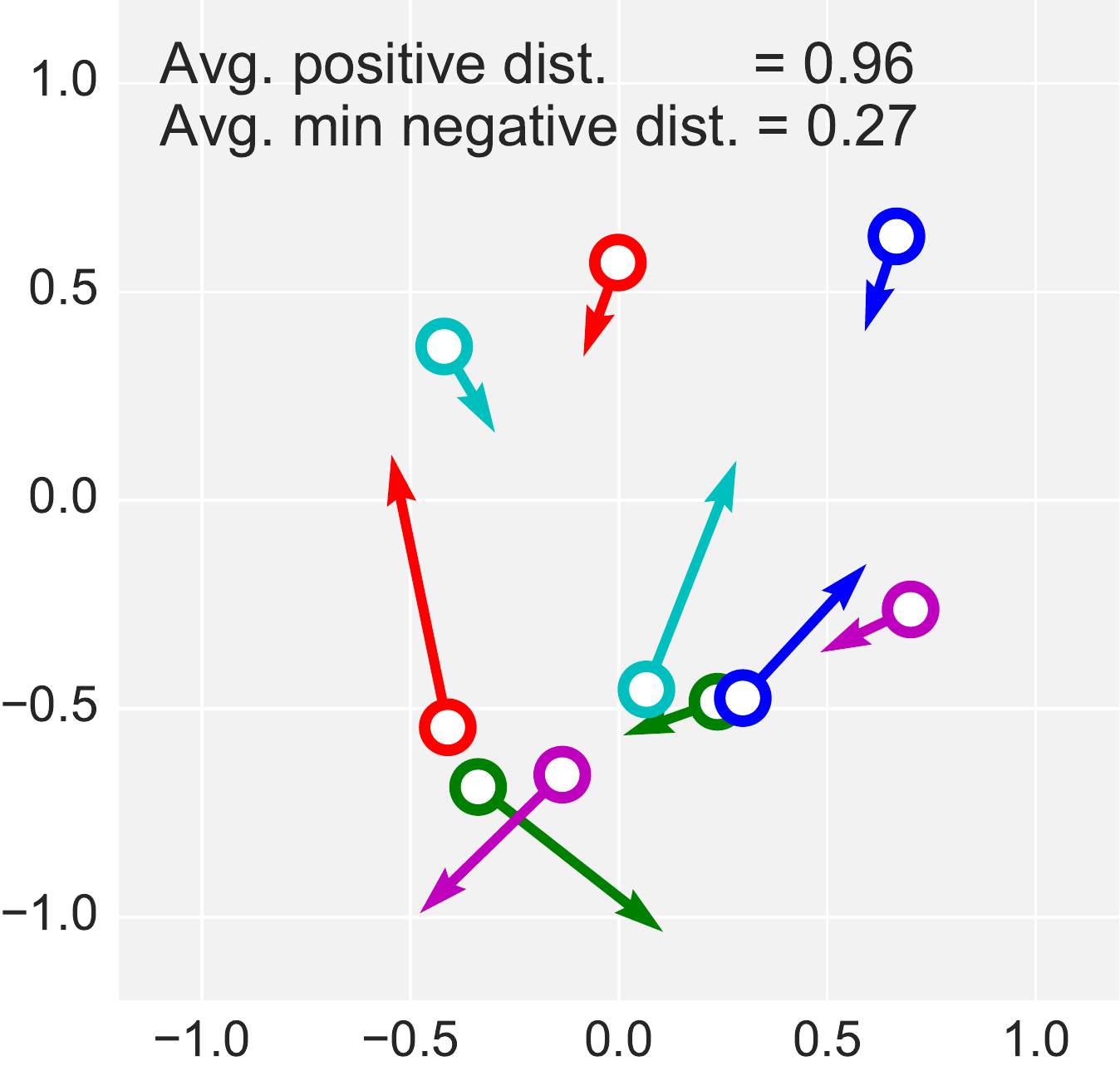}\hfill
 \includegraphics[width=0.30\linewidth]{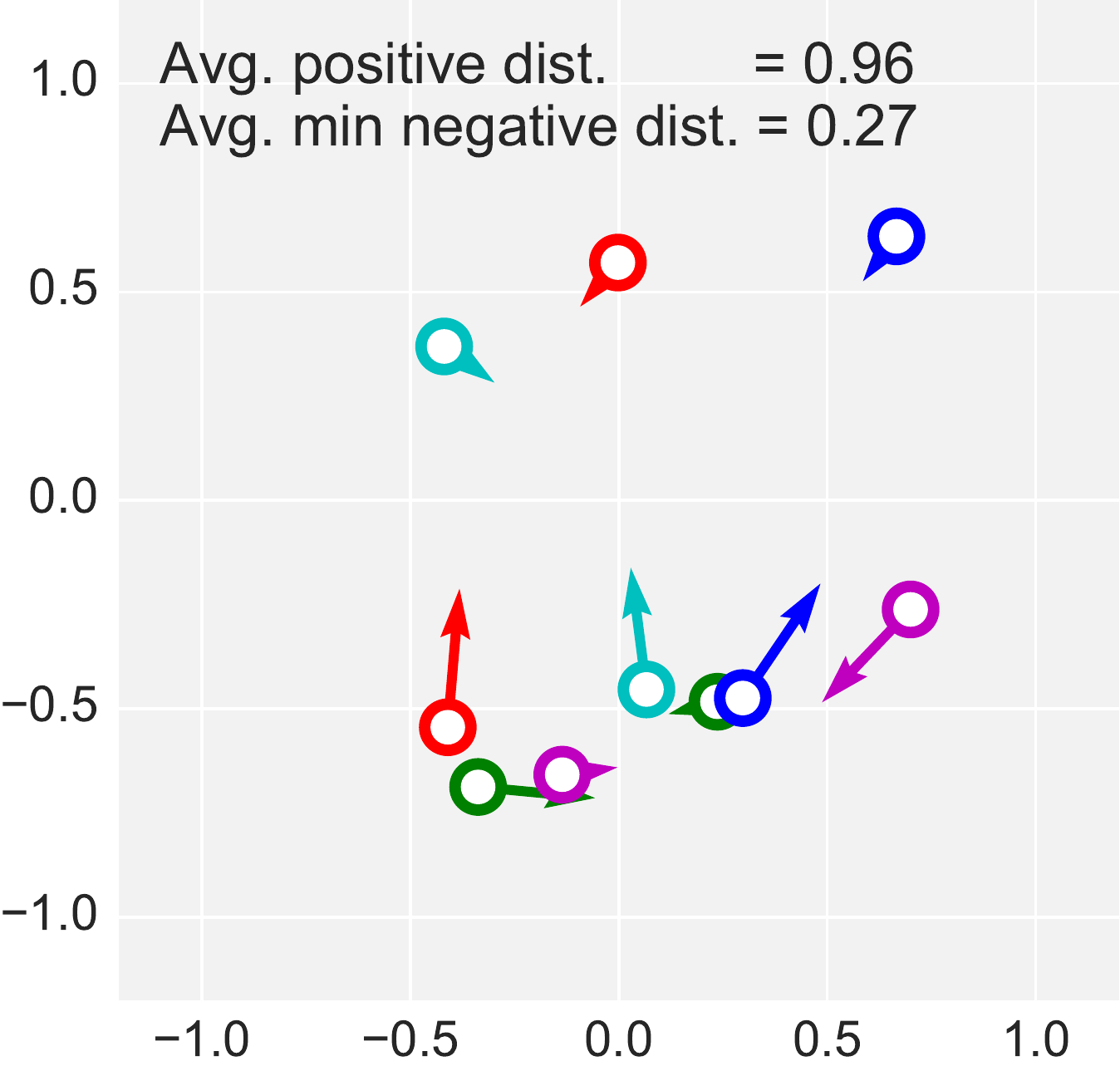}\\
After 150 Adam steps\\
\begin{minipage}[h]{0.30\linewidth}
\center{\includegraphics[width=1\linewidth]{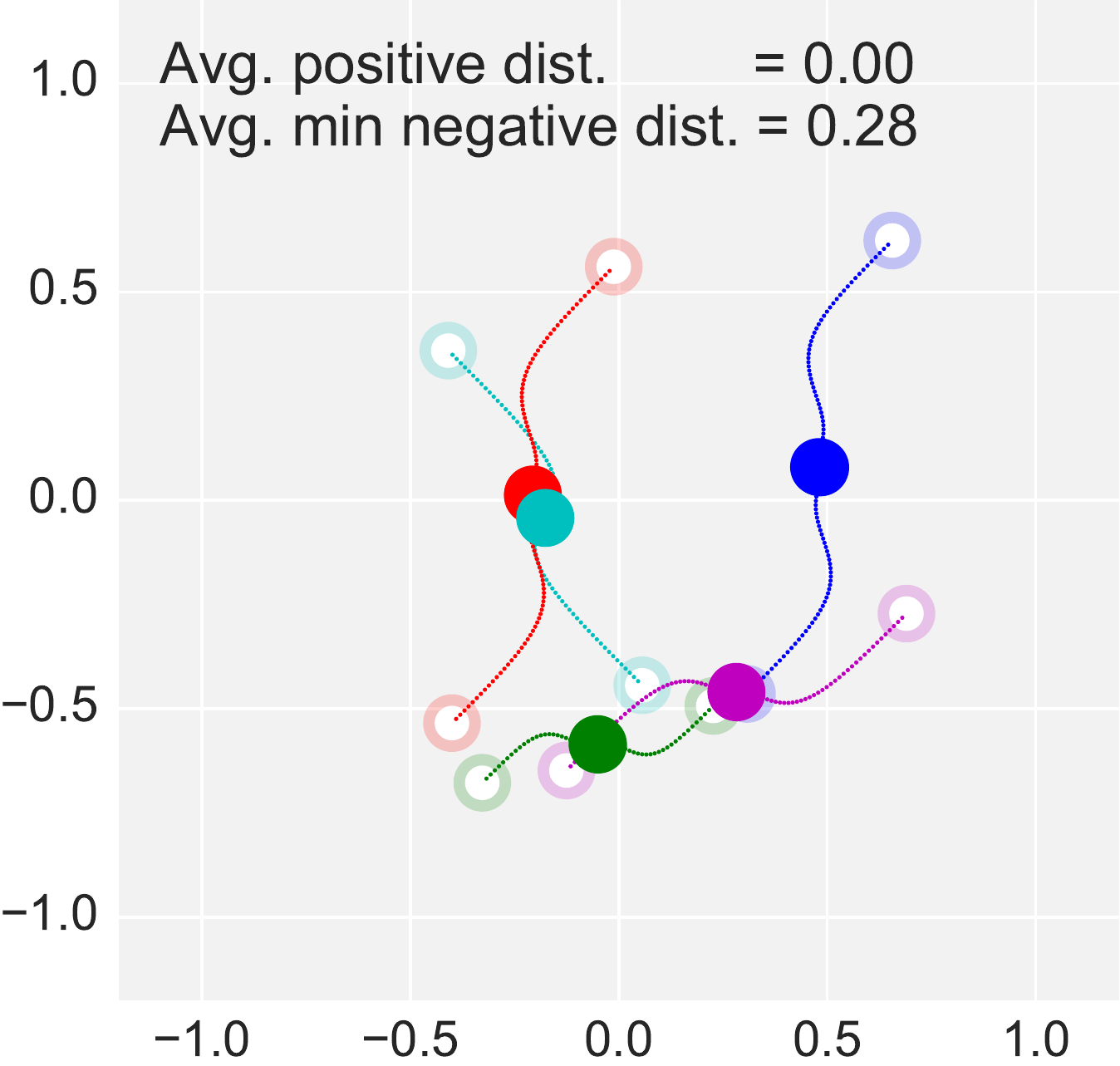}} PosDist loss\\
\end{minipage}
\hfill
\begin{minipage}[h]{0.30\linewidth}
\center{\includegraphics[width=1\linewidth]{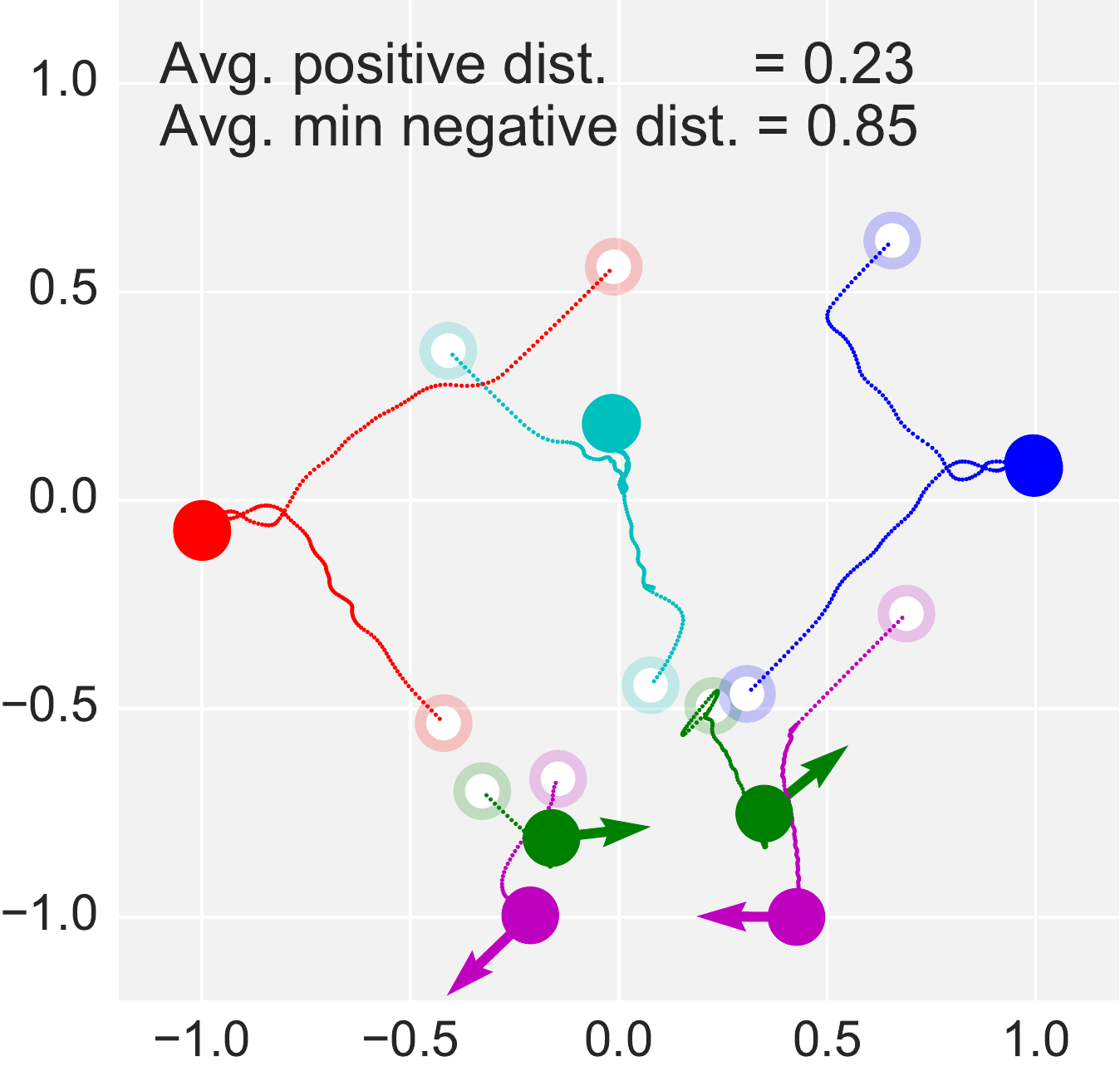}} HardNeg loss\\
\end{minipage}
\hfill
\begin{minipage}[h]{0.30\linewidth}
\center{\includegraphics[width=1\linewidth]{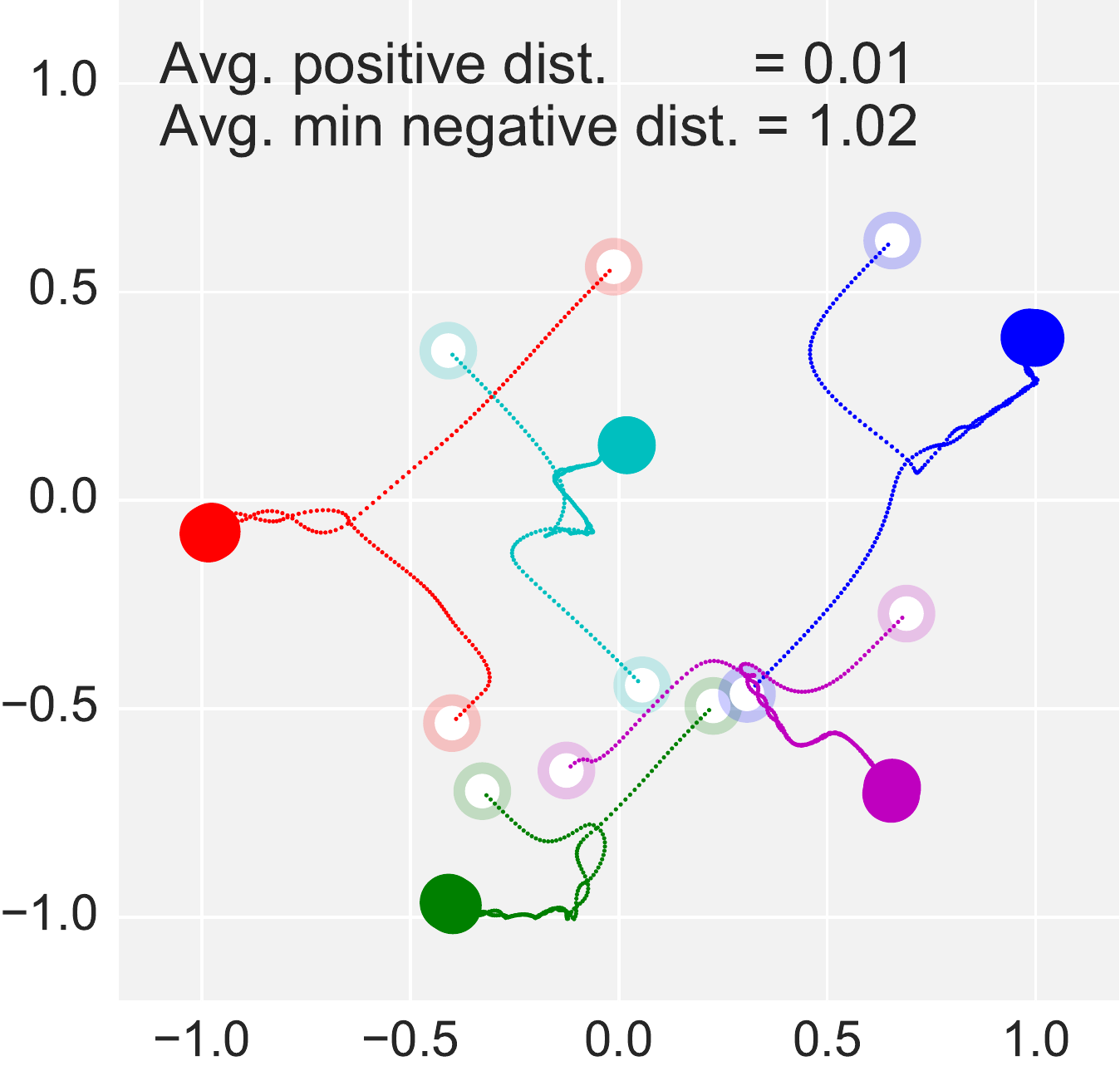}} HardNegC loss\\
\end{minipage}
 \caption{A toy example optimization problem illustrating the proposed hard negative-constant (HardNegC) loss. Five pairs of points, representing 2D descriptors, are generated and the losses are minimized by Adam~\cite{Adam2014}: the positive descriptor distance (PosDist)~\cite{OriNet2016} -- left, the hard negative (HardNeg) margin loss~\cite{HardNet2017} -- center, HardNegC-- right. 
Top row: identical initial positions of five pairs of matching points. Arrows show the gradient direction and relative magnitude. 
Bottom row: points after 150 steps of Adam optimization, trajectories are shown by dots. HardNeg loss has a difficulty with the green and magenta point pairs, because the negative example lies between two positives. Minimization of the positive distance only leads to a small distance to the negative examples. The proposed HardNegC loss first pushes same class points close to each other and then distributes them to increase distance to the negative pairs.}
\label{fig:toy}
\end{figure}
\subsection{The hard negative-constant loss}
\label{sec:loss}
We propose a loss function called hard negative-constant loss (HardNegC). It is based on the hard negative triplet margin loss~\cite{HardNet2017} (HardNeg), but the distance to the hardest (i.e. closest) negative example is treated as constant and the respective derivative of $L$  is set to zero:
\begin{equation}
L = \frac{1}{n}\sum_{i = 1,n}{\max{(0, 1 + d(s_i, \dot{s}_i) - d(s_i, N))}}, \quad
\frac{\partial L}{\partial N} \coloneqq  0,
\label{eq:hardnegc}
\end{equation}
where $d(s_i, \dot{s}_i)$ is the distance between the matching descriptors, $d(s_i, N)$ is a distance to the hardest negative example $N$ in the mini-batch for $i^{th}$ pair.
\begin{equation*}
d(s_i, N) = \min{( \min_{j \ne i}{d(s_i, \dot{s}_j)}, \min_{j\ne i}{d(s_j, \dot{s}_i)})}
\end{equation*}
The difference between the Positive descriptor distance loss (PosDist) used for learning local feature orientation in~\cite{OriNet2016} and the HardNegC and HardNeg losses is shown on a toy example in Figure~\ref{fig:toy}. Five pairs of points in the 2D space are generated and their positions are updated by the Adam optimizer~\cite{Adam2014} for the three loss functions. PosDist converges the first, but the different class points end up near each other, because the distance to the negative classes is not incorporated in the loss. The HardNeg margin loss has trouble when the points from different classes lie between each other. The HardNegC loss behavior first resembles the PosDist loss, bringing positive points together and then distributes them in the space, satisfying the triplet margin criterion.
\subsection{Descriptor losses for shape registration}
\begin{figure}[htb]
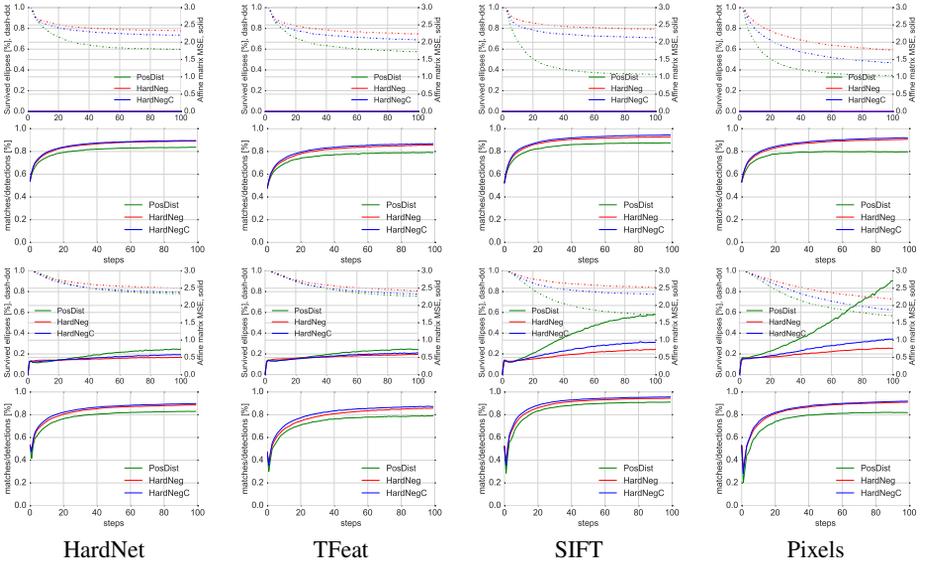

\input fig-direct-opt-ideal-joint
\input fig-direct-opt-ideal
 \caption{Matching score versus geometric repeatability experiment. Affine shape registration by a minimization of descriptor losses of corresponding features. Descriptor losses: green -- L2- descriptor distance (PosDist)~\cite{OriNet2016}, red -- hard triplet margin HardNeg~\cite{HardNet2017}, blue -- proposed HardNegC. Average over HSequences, illumination subset.  
\emph{All features are initially perfectly registered}. 
First two rows: single feature geometry for both images, second two rows: feature geometries are independent in each image.
Top row: geometric error of corresponding features (solid) and percentage of non-collapsed, i.e. elongation $\leq$ 6, features (dashed). Bottom row: the percentage of correct matches.
This experiment shows that even perfectly initially registered feature might not be matched with any of descriptors -- initial matching score is roughly $\approx 30..50\%$. But it is possibly to find measurement region, which offers both discriminativity and repeatability.
 PosDist loss squashes  most of the features, leading to the largest geometrical error. HardNeg loss produces the best results in the number of survived feature and geometrical error. HardNegC performs slightly worse than HardNeg, slightly outperforming it on matching score. However, HardNegC is easier to optimize for AffNet learning -- see Table~\ref{tab:desc-and-loss}.}
\label{fig:direct-opt-ideal}
\end{figure}

Exploring how local feature repeatability is connected with descriptor similarity, we conducted an shape registration experiment (Figure~\ref{fig:direct-opt-ideal}). Hessian features are detected in reference HSequences~\cite{hpatches2017} illumination images and reprojected by (identity) homography to another image in the sequence. Thus, the repeatability is 1 and reprojection error is 0. Then, the local descriptors (HardNet~\cite{HardNet2017}, SIFT~\cite{SIFT2004}, TFeat~\cite{TFeat2016} and raw pixels) are extracted and features are matched by first-to-second-nearest neighbor ratio~\cite{SIFT2004} with threshold 0.8.  This threshold was suggested by Lowe~\cite{SIFT2004} as a good trade-off between false positives and false negatives. For SIFT, 22\% of the geometrically correct correspondences are not the nearest SIFTs and they cannot be matched, regardless of the threshold. In our experiments, the 0.8 threshold worked well for all descriptors and we used it, in line with previous papers, in all experiments.

Notice that for all descriptors, the percentage of correct matches even for the \emph{perfect} geometrical registration is only about 50\%. 

Adam optimizer is used to update affine region $A$ to minimize the descriptor-based losses: PosDist, HardNeg and HardNegC. 
The top two rows show the results for $A$ matrices coupled for both images, bottom -- the descriptor difference optimization is allowed to deform $A$ and $\dot{A}$ in both images independently, which leads to a pair of affine regions that are not in perfect geometric correspondence, yet they are more matchable. Note, that no training of any kind is involved. 

Such descriptor-driven optimization, not maintaining perfect registration, produces a descriptor that is matched successfully up to 90\% of the detections under illumination changes. 

For most of the unmatched regions, the affine shapes become a degenerate lines -- shown in top graphs, and the number of degenerate ellipses is high for PosDist loss; HardNeg and HardNegC perform better.

The bottom row of Figure~\ref{fig:direct-opt-ideal} shows results for experiments where affine shapes pairs are independent in each image. Optimization of  descriptor losses lead to an increase of the geometric error on the affine shape. Error $E$ is defined as the mean square error on A matrix difference:
\begin{equation}
E = \sum_{i=1}^{n} \frac{2(A_i -\dot{A}_i)^2}{\det{A} +\det{\dot{A}}}
\end{equation}
Again, PosDist loss leads to a larger error. CNN-based descriptors, HardNet and TFeat lead to relative small geometric error when reaching matchability plateau, while for SIFT and raw pixels the shapes diverge. Figure~\ref{fig:direct-opt-random} shows the case when the initialized shapes include a small amount of the reprojection error. 
\begin{figure}[htb]
\input fig-direct-opt-noise
\caption{Minimization of descriptor loss by optimization of affine parameters of corresponding features. Average over HPatchesSeq, illumination subset. Top row: geometric error of corresponding features (full line)  and percentage of non-collapsed, \ie elongation $\leq$ 6, features (dashed line). Bottom row: the fraction correct matches.
 All features initially have the same medium amount of reprojection noise.  Left to right: HardNet, SIFT, TFeat, mean-normalized pixels descriptors.}
 \label{fig:direct-opt-random}
 \end{figure}
\begin{figure}[t]
\centering
 \includegraphics[width=0.99\linewidth]{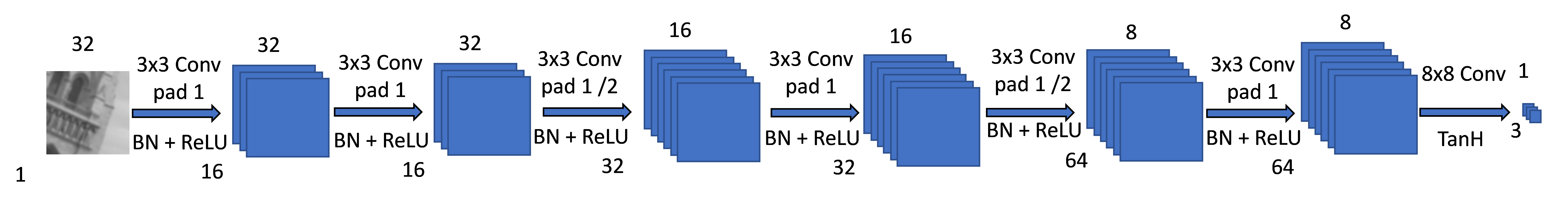}\\
 \caption{AffNet. Feature map spatial size -- top, \# channels -- bottom. /2 stands for stride 2.}
 \label{fig:architecture}
\end{figure}
\begin{figure}[htb]
\centering
\includegraphics[width=0.95\linewidth]{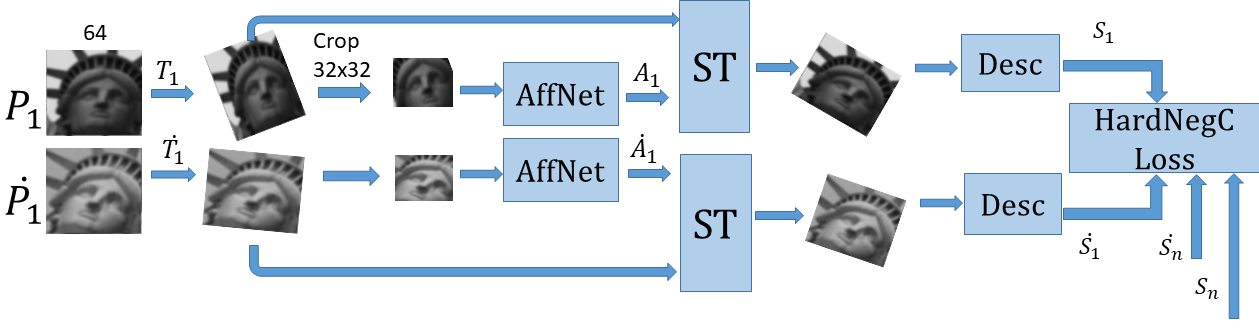}
 \caption{AffNet training. Corresponding patches undergo random affine transformation $T_i, \dot{T}_i$, are cropped and fed into AffNet, which outputs affine transformation $A_i, \dot{A}_i$ to an unknown canonical shape. ST -- the spatial transformer warps the patch into an estimated canonical shape. The patch is described by a differentiable CNN descriptor. $n \times n$ descriptor distance matrix is calculated and used to form triplets, according to the HardNegC loss.}
\label{fig:training-scheme}
\end{figure}
\subsection{AffNet training procedure}
The main blocks of the proposed training procedure are shown in Figure~\ref{fig:training-scheme}. 
First, a batch of matching patch pairs $(P_i, \dot{P}_i)_{i=1..n}$ is generated, where $P_i$ and $\dot{P}_i$ correspond to the same point on a 3D surface. Rotation and skew transformation matrices $(T_i, T_i')$ are randomly and independently generated. The patches $P_i$ and $\dot{P}_i$ are warped by  $(T_i, \dot{P}_i)$ respectively into $A$-transformed patches. Then, a $32\times 32$ center patch is cropped and a pair of transformed patches is fed into the convolutional neural network AffNet, which predicts a pair of affine transformations $A_i$, $\dot{P}_i$, that are applied to the $T_i$-transformed patches via spatial transformers ST~\cite{SpatialTransformers2015}. 

Thus, geometrically normalized patches are cropped to $32\times 32$ pixels and fed into the descriptor network, e.g. HardNet, SIFT or raw patch pixels, obtaining descriptors $(s_i, \dot{s}_i)$. 
Descriptors $(s_i, \dot{s}_i)$ are then used to form triplets by the procedure proposed in~\cite{HardNet2017}, followed by our newly proposed hard negative-constant loss (Eq.~\ref{eq:hardnegc}).

More formally, we are finding affine transformation model parameters $\theta$ such that estimated affine transformation $A$ minimizes descriptor HardNegC loss:
\begin{equation}
A(\theta | (P,\dot{P})) = \argmin_{\theta}{L(s,\dot{s})}
\end{equation}
\subsection{Training dataset and data preprocessing}
UBC Phototour~\cite{Brown2007} dataset is used for training. It consists of three subsets: \emph{Liberty}, \emph{Notre Dame} and \emph{Yosemite} with about 2 $\times$ 400k normalized 64x64 patches in each, detected by DoG and Harris detectors. Patches are verified by 3D reconstruction model. We randomly sample 10M pairs for training.

Although positive point corresponds to roughly the same point on the 3D surface, they are not perfectly aligned, having position, scale, rotation and affine noise.
We have randomly generated affine transformations, which consist in random rotation -- tied for pair of corresponding patches, and anisotropic scaling $t$ in random direction by magnitude $t_m$, which is gradually increased during the training from the initial value of 3 to 5.8 at the middle of the training. The tilt is uniformly sampled from range $[0,t_m]$. 
\subsection{Implementation details}
The CNN architecture is adopted from HardNet\cite{HardNet2017}, see Fig.~\ref{fig:architecture}, 
with the number of channels in all layers reduced  2x and the last 128D output  replaced by  a 3D output predicting ellipse shape. 
The network formula is 16C3-16C3-32C3/2-32C3-64C3/2-64C3-3C8, where 32C3/2 stands for 3x3 kernel with 32 filters and stride 2. Zero-padding is applied in all convolutional layers to preserve the size, except the last one. BatchNorm~\cite{BatchNorm2015} layer followed by ReLU~\cite{Nair2010RectifiedLinearUnits} is added after each convolutional layer, except the last one, which is followed by hyperbolic tangent activation.
Dropout~\cite{Dropout2014} with 0.25 rate is applied before the last convolution layer. Grayscale input patches $32 \times 32$ pixels are normalized by subtracting the per-patch mean and dividing by the per-patch standard deviation. 

Optimization is done by SGD with learning rate 0.005, momentum 0.9, weight decay 0.0001. The learning rate decayed linearly~\cite{Systematic2017} to zero within 20 epochs. The training was done with PyTorch~\cite{pytorch} and took 24 hours on Titan X GPU; the bottleneck is the data augmentation procedure. The inference time is 0.1~ms per patch on Titan X, including patch sampling done on CPU and Baumberg iteration -- 0.05~ms per patch on CPU.
\section{Empirical evaluation}
\label{sec:exp}
\begin{figure}[tb]
\centering
  \includegraphics[width=0.24\linewidth]{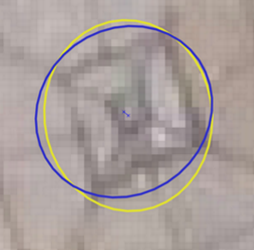}
   \includegraphics[width=0.24\linewidth]{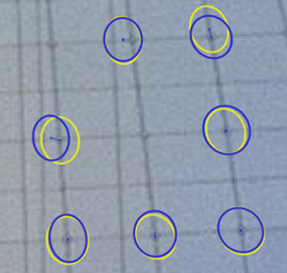}
  \includegraphics[width=0.24\linewidth]{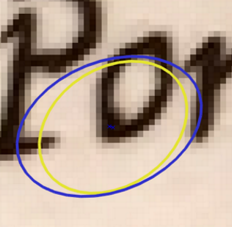}
   \includegraphics[width=0.24\linewidth]{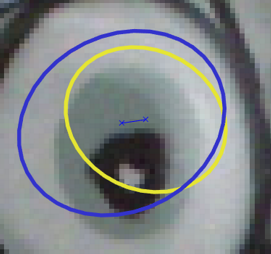}\\
     \includegraphics[width=0.24\linewidth]{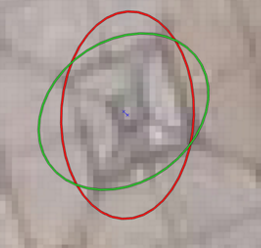}
   \includegraphics[width=0.24\linewidth]{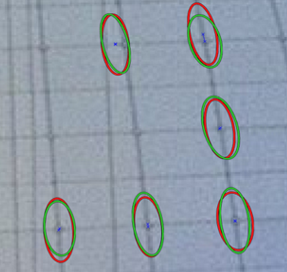}
  \includegraphics[width=0.24\linewidth]{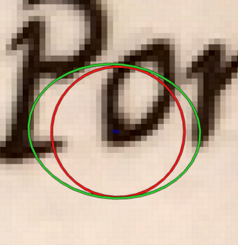}
   \includegraphics[width=0.24\linewidth]{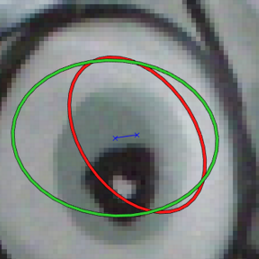}\\
 \caption{ AffNet (top) and Baumberg (bottom) estimated affine shape. One ellipse is detected in the reference image, the other is a reprojected closest match from the second image. Baumberg ellipses tend to be more elongated, average axis ratio is 1.99 vs. 1.63 for AffNet, median: Baumberg 1.72 vs 1.39 AffNet. The statistics are calculated over 16M features on Oxford5k.}
 \label{fig:patch-examples}
\end{figure}
\subsection{Loss functions and descriptors for learning measurement region}
We trained different versions of the AffNet and orientation networks, with different combinations affine transformation parameterizations and descriptors with the procedure described above. 
The results of the comparison based on the number of correct matches (reprojection error $\leq$ 3 pixel) on the hardest pair for each of the 116 sequences from the HSequences~\cite{hpatches2017} dataset are shown in Tables~\ref{tab:desc-and-loss},\ref{tab:aff-shape-parametrization}.

The proposed HardNetC loss is the only loss function with no "not converged" results. In the case of convergence, all tested descriptors and loss functions lead to comparable performance, unlike registration experiments in the previous section. We believe it is because now the CNN always outputs the same affine transformation for a patch, unlike in the previous experiment, where repeated features may end up with different shapes.

Affine transformation parameterizations are compared in Table~\ref{tab:aff-shape-parametrization}. All attempts to learn affine shape and orientation jointly in one network fail completely, or perform significantly worse than the two-stage procedure, when affine shape is learned first and orientation is estimated on an affine-shape-normalized patch. 
Learning residual shape $A''$ (Eq.~\ref{eq:A''}) leads to the best results overall. Note, that such parameterization does not contain enough parameters to include feature orientation, thus "joint" learning is not possible. Slightly worse performance is obtained by using an identity matrix prior for learnable biases in the output layer. 
\begin{table}[htb]
\ra{1}
\centering
\caption{Learning the affine transform: loss functions and descriptor comparison. The median of average number of correct matches on the HSequences~\cite{hpatches2017} hardest image pairs 1-6 for the Hessian detector and the HardNet descriptor. The match considered correct for reprojection error $\leq$ 3 pixels. Affine shape is parametrized as in Eq.~\ref{eq:A''}. n/c -- did not converge.}
\label{tab:desc-and-loss}
\setlength{\tabcolsep}{2mm}
\begin{tabular}{cccc}
\toprule
Training descriptor/loss & PosDist& HardNeg & HardNegC  \\
\cmidrule(r){1-4}
\multicolumn{4}{c}{Affine shape}\\
\cmidrule(r){1-4}
SIFT & n/c & 385 &  386 \\
HardNet & n/c & n/c &  \textbf{388} \\
\cmidrule(r){1-4}
Baumberg~\cite{Baumberg2000} &\multicolumn{3}{c}{298}  \\
\cmidrule(r){1-4}
\multicolumn{4}{c}{Orientation}\\
\cmidrule(r){1-4}
SIFT &  \textbf{387} & 379 &  382 \\
HardNet & 386 & 383 & 380 \\
\cmidrule(r){1-4}
Dominant orientation~\cite{SIFT2004} &\multicolumn{3}{c}{339}  \\
\bottomrule
\end{tabular}
\end{table}
\begin{table}[htb]
\centering
\caption{Learning the affine transform: parameterization comparison. The average number of correct matches on the HPatchesSeq~\cite{hpatches2017} hardest image pairs 1-6 for the Hessian detector and the HardNet descriptor.
Cases compared, affine shape combined with the de-facto handcrafted standard dominant orientation, affine shape and orientation learnt separately or jointly. The match considered correct for reprojection error $\leq$ 3 pixels. The HardNegC loss and HardNet descriptor used for learning. n/c -- did not converge.}
\label{tab:aff-shape-parametrization}
\setlength{\tabcolsep}{2mm}
\begin{tabular}{lcllccc}
\toprule
&&&& \multicolumn{3}{c}{Orientation} \\
\cmidrule(r){5-7}
& & Estimated& biases & \multicolumn{2}{c}{Learned} & Dominant \\ 
\cmidrule(r){5-6}
Eq. &Matrix& parameters &  init & jointly & separately & gradient~\cite{SIFT2004} \\ 
\cmidrule(r){1-7}
(\ref{eq:A})&$A$ & $(a_{11},a_{12}, a_{21},a_{22})$ & 0 & n/c & n/c  & n/c \\
(\ref{eq:A})&$A$  & $(a_{11},a_{12}, a_{21},a_{22})$ & 1 & n/c & 360  & 320 \\
\cmidrule(r){1-7}
(\ref{eq:A'}) &$A'$,& $(a'_{11},0 , a'_{21},a'_{22})$, & 1 & 250 & 327  & 286 \\
&  $R(\alpha)$  & $(\sin{\alpha}, \cos{\alpha}$) & & & \\
\cmidrule(r){1-7}
(\ref{eq:A''})& $A''$ & $(a''_{11},a''_{21},a''_{22})$&  1 & - &  370 & 340  \\
(\ref{eq:A''})&$A''$ & $(1 + a''_{11},a''_{21}, 1 + a''_{22})$&  0 & - &  \textbf{388} & 349  \\
\bottomrule
\end{tabular}
\end{table}
\begin{figure}[htb]
\centering
 \includegraphics[width=0.24\linewidth]{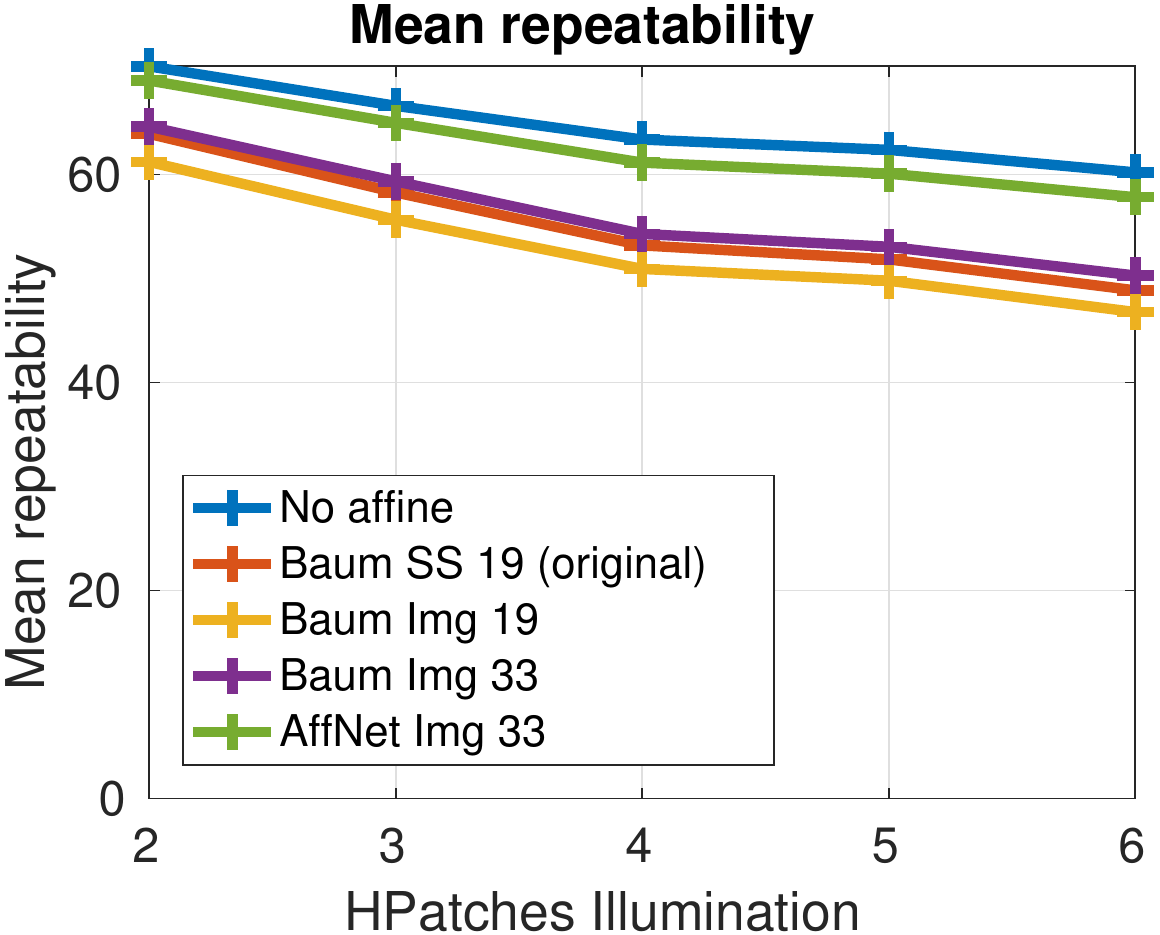}
 \includegraphics[width=0.24\linewidth]{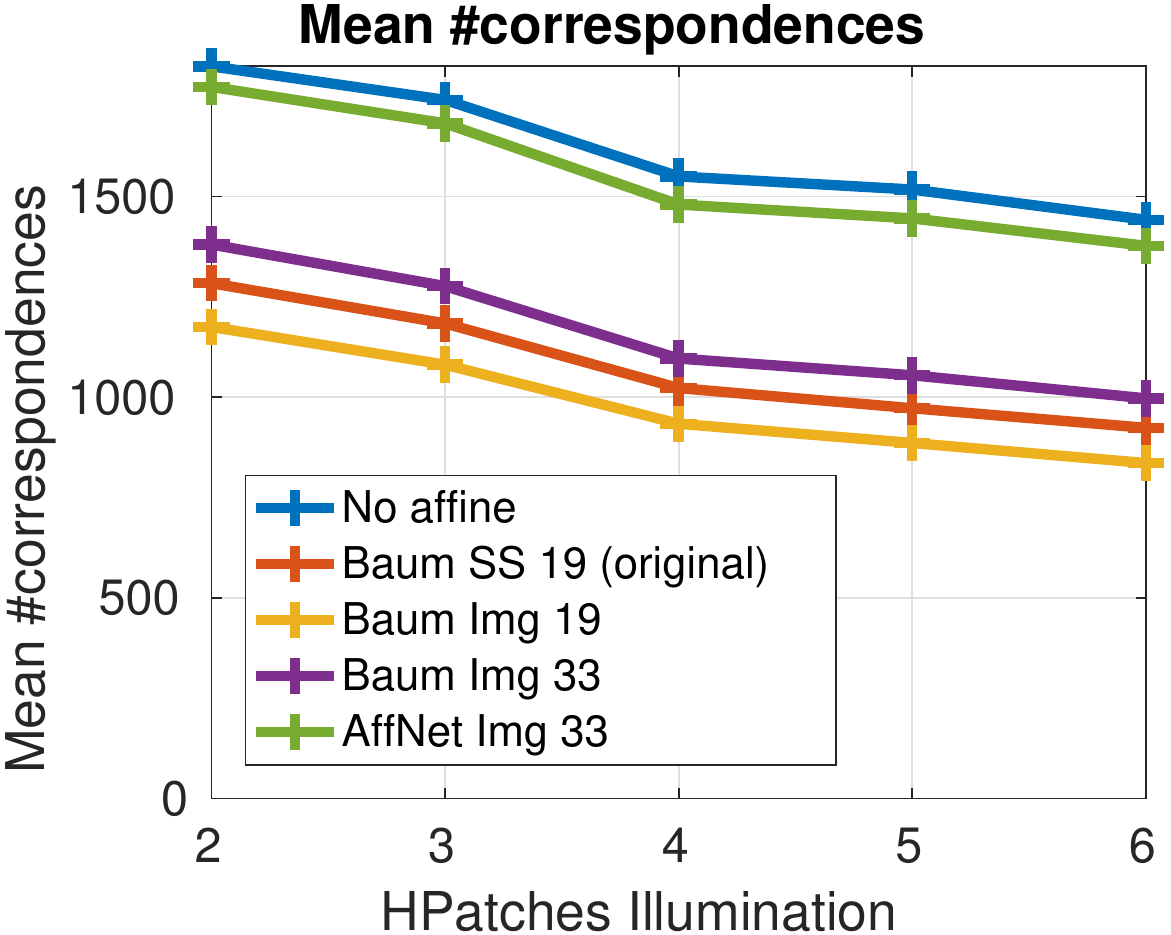}
 \includegraphics[width=0.24\linewidth]{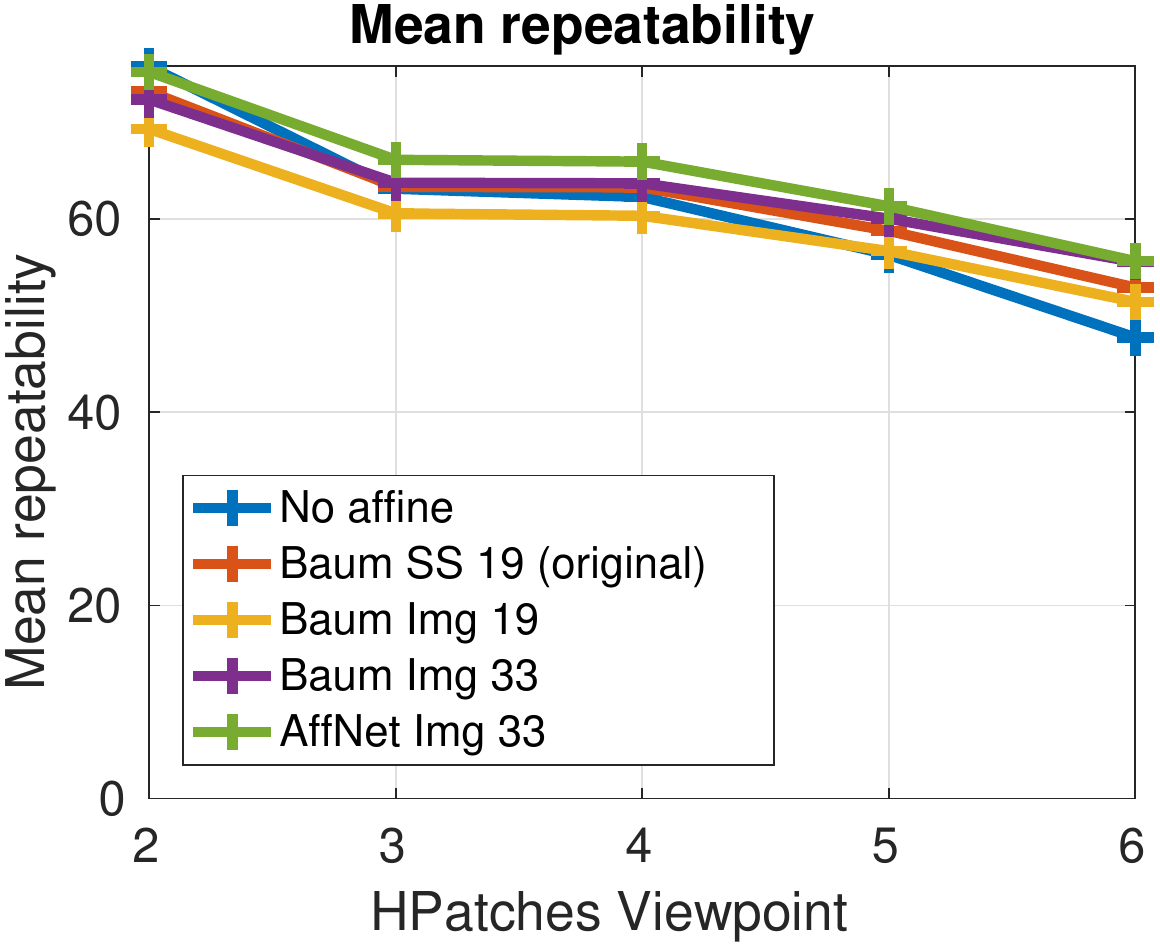}
 \includegraphics[width=0.24\linewidth]{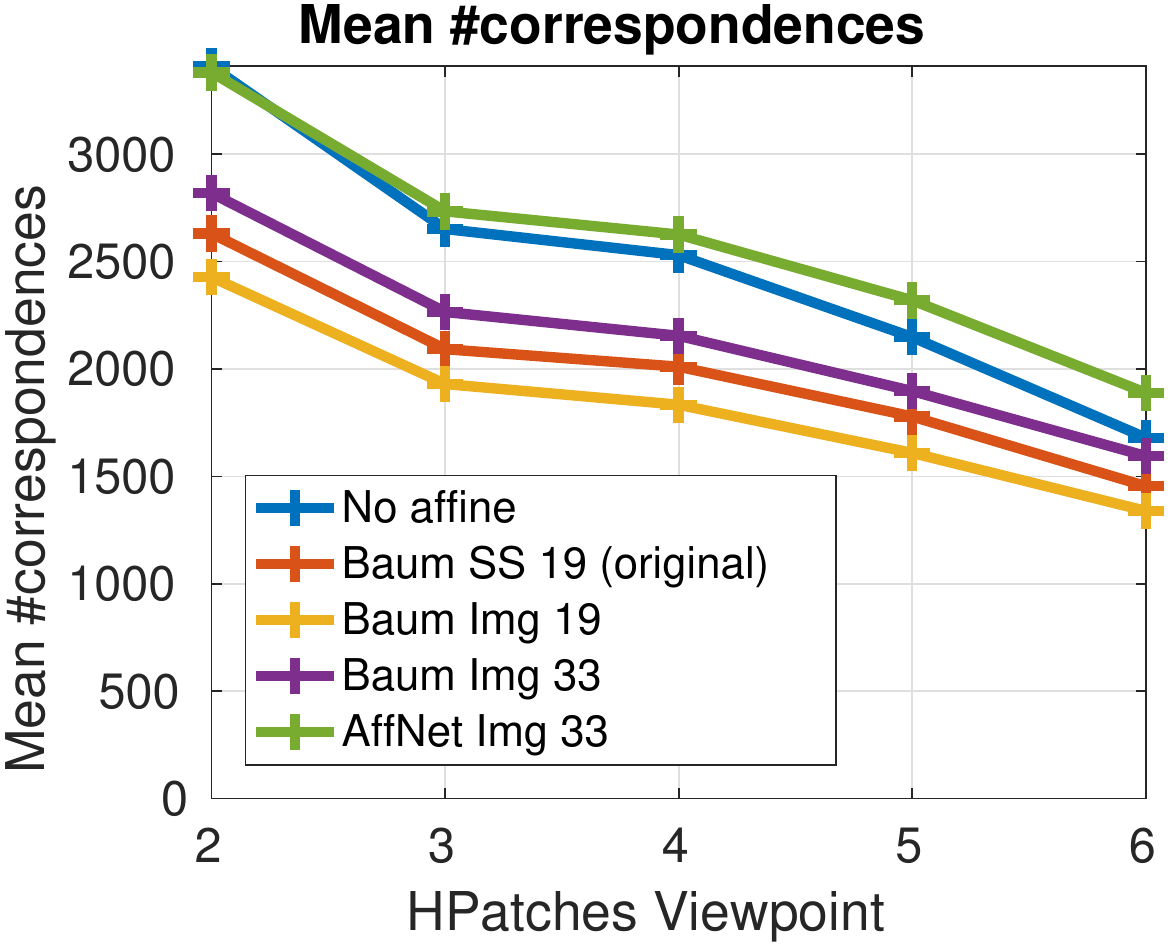}\\
 \includegraphics[width=0.24\linewidth]{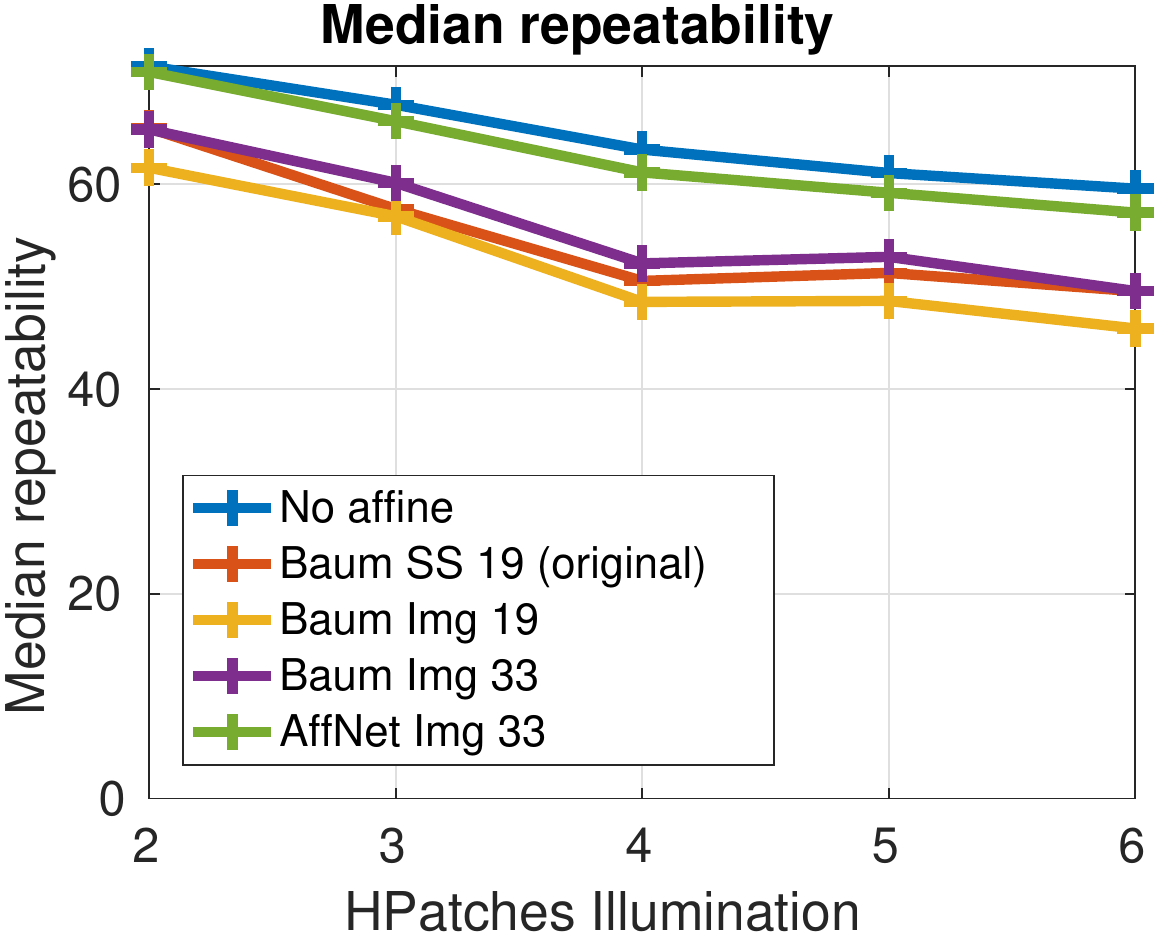}
 \includegraphics[width=0.24\linewidth]{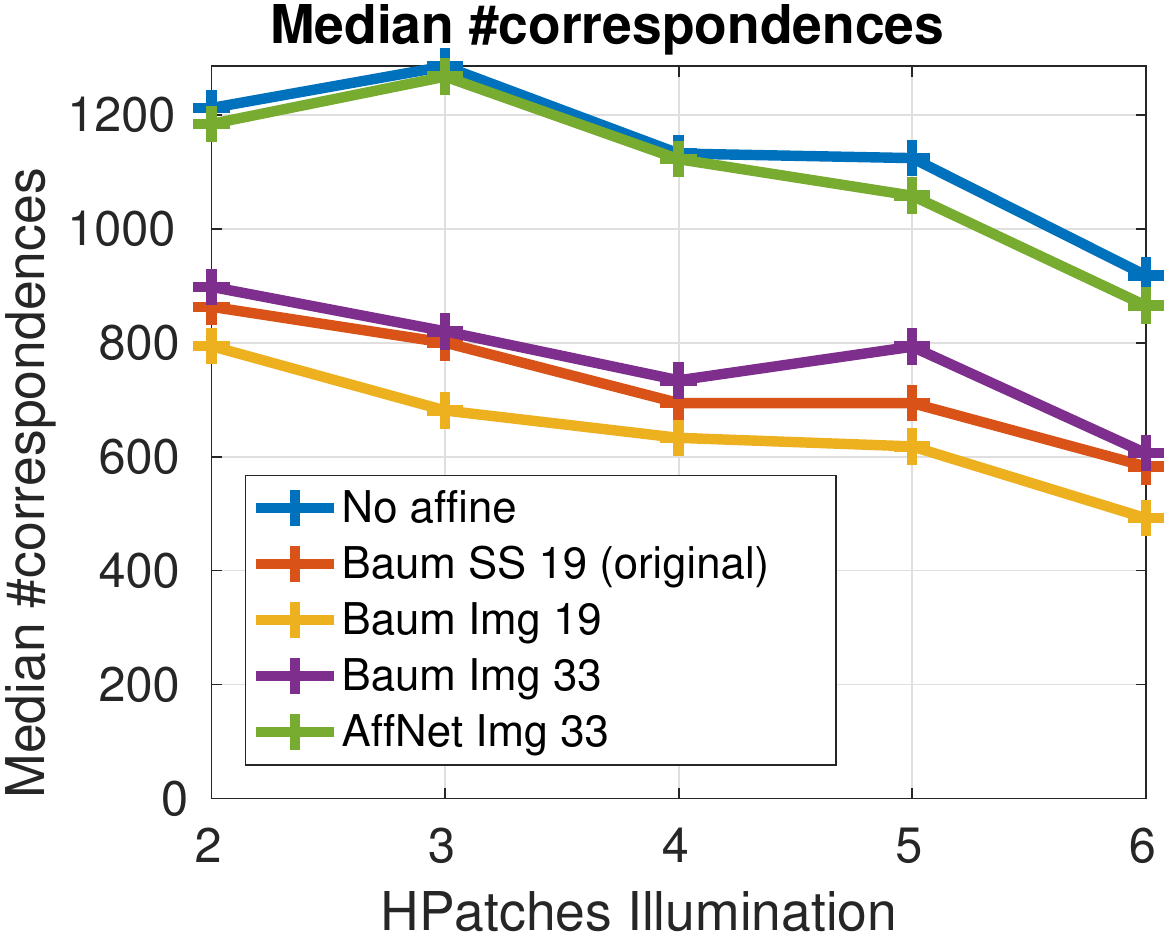}
 \includegraphics[width=0.24\linewidth]{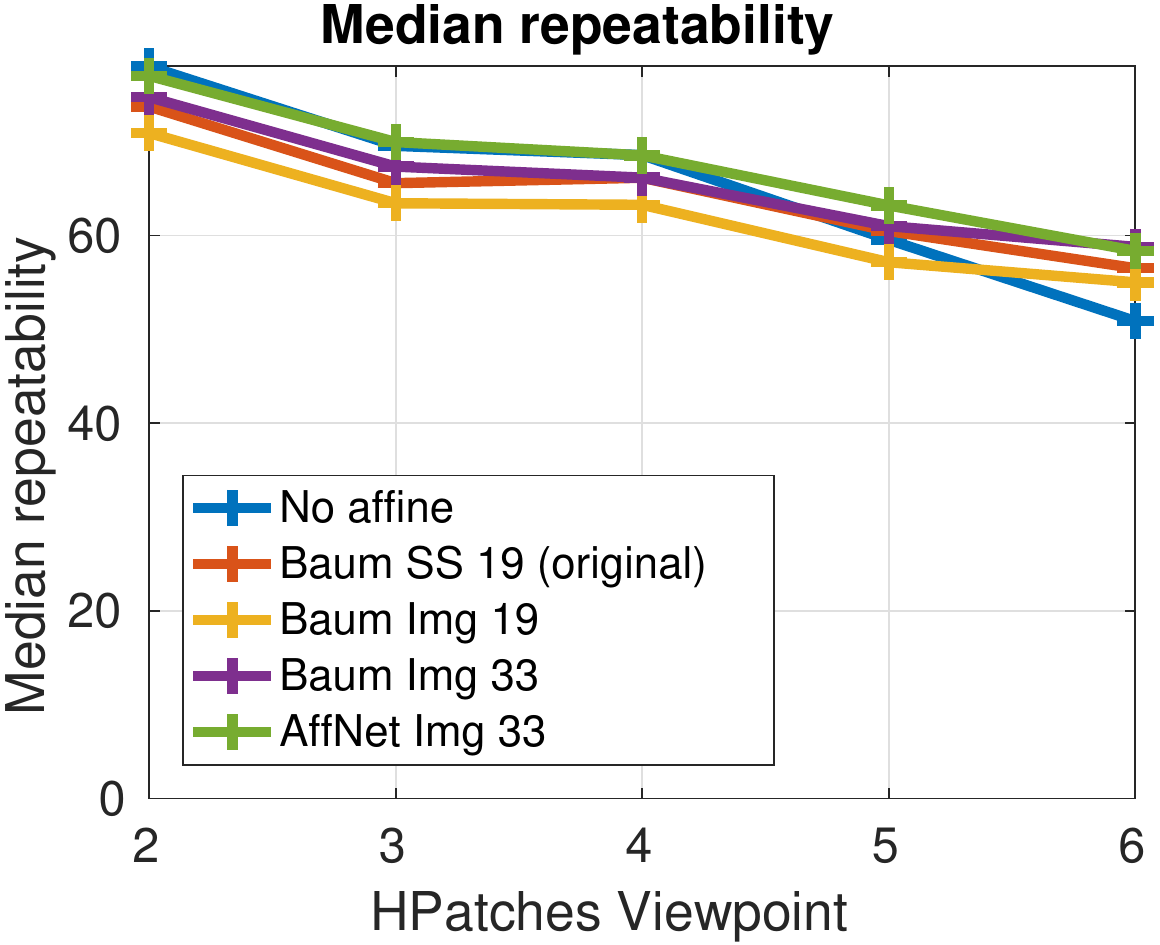}
 \includegraphics[width=0.24\linewidth]{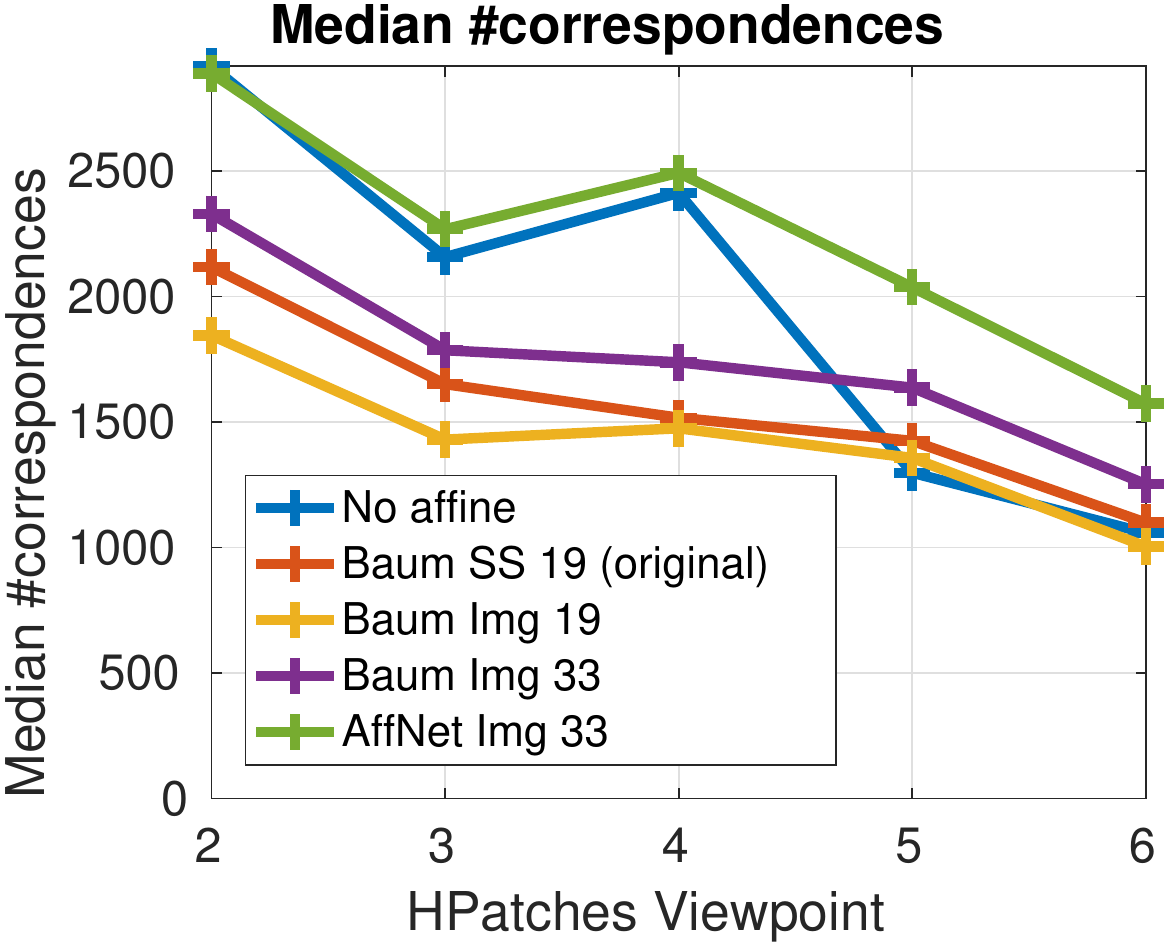}\\
 \caption{Repeatability and the number of correspondences (mean top, median bottom row) on the HSequences~\cite{hpatches2017}. AffNet is compared with the de facto standard Baumberg iteration~\cite{Baumberg2000} according to the Mikolajczyk protocol~\cite{Mikolajczyk2005}. Left -- images with illumination differences, right -- with viewpoint and scale changes. SS -- patch is sampled from the scale-space pyramid at the level of the detection, image -- from the original image; 19 and 33 -- patch sizes. Hessian-Affine is from\cite{Perdoch-CVPR2009efficient}. For illumination subset, performance of Hessian with no adaptation is an upper bound, and AffNet performs close to it.}
 \label{fig:repeatability}
\end{figure}
\subsection{Repeatability}
\label{exp:rep}
Repeatability of affine detectors: Hessian detector + affine shape estimator was benchmarked, following classical work by Mikolajczyk~\etal~\cite{Mikolajczyk2005}, but on recently introduced larger HSequences~\cite{hpatches2017} dataset by VLBenchmarks toolbox~\cite{lenc2012vlbenchmarks}.

HSequences consists of two subsets. \textit{Illumination} part contains 57 image sixplets with illumination changes, both natural and artificial. There is no difference is viewpoint in this subset, geometrical relation between images in sixplets is identity.Second part is \textit{Viewpoint}, where 59 image sixplets vary in scale, rotation, but mostly in horizontal tilt. The average viewpoint change is a bit smaller than in well-known \emph{graffiti} sequence from Oxford-Affine dataset~\cite{Mikolajczyk2005}.

Local features are detected in pairs of images, reprojected by ground truth homography to the reference image and closest reprojected region is found for each region from reference image. The correspondence is considered correct, when overlap error of the pair is less than 40\%. The repeatability score for a given pair of images is a ratio between number of correct correspondences and the smaller number of detected regions in common part of scene among two images. 

Results are shown in Figure~\ref{fig:repeatability}. Original affine shape estimation procedure, implemented in~\cite{Perdoch-CVPR2009efficient} is denoted Baum SS 19, as $19\times 19$ patches are sampled from scale space. AffNet takes $32\times 32$ patches, which are sampled from original image. So for fair comparison, we also tested Baum versions, where patches are sampled from original image, with 19 and 33 pixels patch size.

AffNet slightly outperforms all the variants of Baumberg procedure for images with viewpoint change in terms of repeatability and more significant -- in number of correspondences.
The difference is even bigger for them image with illumination change only, where AffNet performs almost the same as plain Hessian, which is upper bound here, as this part of dataset has no viewpoint changes. 

We have also tested AffNet with other detectors on the Viewpoint subset of the HPatches. The repeatabilities are the following (no affine adaptation/Baumberg/AffNet): DoG: 0.46/0.51/0.52, Harris: 0.41/0.44/0.47, Hessian: 0.47/0.52/0.56
The proposed methods outperforms the standard (Baumberg) for all detectors. 

One reason for such difference is the feature-rejection strategy.
Baumberg iterative procedure rejects feature in one of three cases. First, elongated ellipses with long-to-short axis ratio more than six are rejected.
Second, features touching boundary of the image are rejected. This is true for the AffNet post-processing procedure as well, but AffNet produces less elongated shapes: average axis ratio on Oxford5k 16M features is 1.63 vs. 1.99 for Baumberg. Both cases happen less often for AffNet, increasing the number of surviving features by 25\%. We compared performance of the Baumberg vs. AffNet on the same number of features in Section~\ref{exp:retrieval}. 
Finally, features whose shape did not converge within sixteen iteration are removed. This is quite rare, it happens in approximately 1\% cases. 
Example of shapes estimated by AffNet and the Baumberg procedure are shown in Fig.~\ref{fig:patch-examples}. 
\input wxbstable
\subsection{Wide baseline stereo}
\label{exp:wbs}
We conducted an experiment on wide baseline stereo, following local feature detector benchmark protocol, defined in~\cite{WXBS2015} on the set of two-view matching datasets~\cite{Hauagge2012,Yang2007,Zitnick2011,Fernando2015}. The local features are detected by benchmarked detector, described by HardNet++~\cite{HardNet2017} and HalfRootSIFT~\cite{HalfSIFT2007} and geometrically verified by RANSAC~\cite{Lebeda2012}. Two following metrics are reported: the number of successfully matched image pairs and average number of correct inliers per matched pair. We have replaced original affine shape estimator in Hessian-Affine with AffNet in Hessian and Adaptive threshold Hessian (AdHess) 

The results are shown in Table~\ref{tab:wxbs-table}. AffNet outperforms Baumberg in both number of registered image pairs and/or number of correct inliers in all datasets, including painting-to-photo pairs in SymB~\cite{Hauagge2012} and multimodal pairs in GDB~\cite{Yang2007}, despite it was not trained for that domains.

The total runtimes per image are the following (average for 800x600 images). Baseline HesAff + dominant gradient orientation + SIFT: no CNN components – 0.4 sec. HesAffNet (CNN) + dominant gradient orientation + SIFT – 0.8s, 3 CNN components: HesAffNet + OriNet + HardNet – 1.2 s. Now the data is naively transferred from CPU to GPU and back each of the stages, which generates the major bottleneck.

\subsection{Image retrieval}
\label{exp:retrieval}
\begin{table*}[tb]
\ra{1}
\centering
\caption{Performance (mAP) evaluation of the bag-of-words (BoW) image retrieval on the Oxford5k and Paris6k benchmarks. Vocabulary consisting of 1M visual words is learned on independent dataset: Oxford5k vocabulary for Paris6k evaluation and \emph{vice versa}. SV: spatial verification. QE($t$): query expansion with $t$ inliers threshold. The best results are in \textbf{bold}.}
\label{tab:ox5kpar6k}
\setlength{\tabcolsep}{1mm}
\resizebox{\textwidth}{!}{%
\begin{tabular}{lc@{\hskip 5mm}c@{\hskip 3mm}ccc@{\hskip 5mm}c@{\hskip 3mm}cc}
\toprule
 & \multicolumn{4}{c}{Oxford5k} & \multicolumn{4}{c}{Paris6k} \\
 \cmidrule(r){2-5} \cmidrule(r){6-9}
Detector--Descriptor & \tiny{BoW} & \tiny{+SV} & \tiny{+SV+QE(15)} & \tiny{+SV+QE(8)} & \tiny{BoW} & \tiny{+SV} & \tiny{+SV+QE(15)} & \tiny{+SV+QE(8)} \\
\cmidrule(r){1-9}
HesAff--RootSIFT~\cite{RootSIFT2012} & 55.1	& 63.0 & 78.4 & 80.1 & 59.3 & 63.7 & 76.4 & 77.4 \\
HesAffNet--RootSIFT & 61.6 & 72.8 & 86.5 & 88.0 & 63.5 & 71.2 & 81.7 & 83.5 \\
\cmidrule(r){1-9}
HesAff--TFeat-M*~\cite{TFeat2016} & 46.7 & 55.6 & 72.2 & 73.8 & 43.8 & 51.8 & 65.3 & 69.7\\
HesAffNet--TFeat-M* & 45.5 & 57.3 & 75.2 & 77.5 & 50.6 & 58.1 & 72.0 & 74.8\\
\cmidrule(r){1-9}
HesAff--HardNet++~\cite{HardNet2017} & 60.8 & 69.6 & 84.5 & 85.1 & 65.0 & 70.3 & 79.1 & 79.9\\
HesAffNetLess--HardNet++ & 64.3 & 73.3 & 86.1 & 87.3 & 62.0 & 68.7 & 79.1 & 79.2\\
HesAffNet--HardNet++ & \textbf{68.3} & \textbf{77.8} & \textbf{89.0} & \textbf{91.1} & \textbf{65.7} & \textbf{73.4} & \textbf{83.3} & \textbf{83.3}\\
\bottomrule
\end{tabular}%
}
\end{table*}
\begin{table*}[tb] 
\ra{1}
\centering
\caption{Performance (mAP) comparison with the state-of-the-art in local feature-based image retrieval. Vocabulary is learned on independent dataset: Oxford5k vocabulary for Paris6k evaluation and \emph{vice versa}. All results are with spatial verification and query expansion. VS: vocabulary size. SA: single assignment. MA: multiple assignments. The best results are in \textbf{bold}.}
\label{tab:ox5kpar6kSOTA}
\setlength{\tabcolsep}{3mm}
\begin{tabular}{lccccc}
\toprule
 & & \multicolumn{2}{c}{Oxford5k} & \multicolumn{2}{c}{Paris6k} \\
 \cmidrule(r){3-4} \cmidrule(r){5-6}
Method & VS & SA & MA & SA & MA\\
\cmidrule(r){1-6}
HesAff--SIFT--BoW-fVocab~\cite{Mikulik-IJCV2013FineVocab} & 16M & 74.0 & 84.9 & 73.6 & 82.4 \\
HesAff--RootSIFT--HQE~\cite{Tolias-PR2014HQE} & 65k & 85.3 & 88.0 & 81.3 & 82.8 \\
HesAff--HardNet++--HQE~\cite{HardNet2017} & 65k & 86.8 & 88.3 & 82.8 & 84.9 \\
\cmidrule(r){1-6}
HesAffNet--HardNet++--HQE & 65k & \textbf{87.9} & \textbf{89.5} & \textbf{84.2} & \textbf{85.9} \\
\bottomrule
\end{tabular}
\end{table*}
\begin{table*}[tb] 
\ra{1.0}
\centering
\caption{Performance (mAP, mP@10) comparison with the state-of-the-art in image retrieval on the R-Oxford and R-Paris benchmarks~\cite{revisitop}. SV: spatial verification. HQE: hamming query expansion. $\alpha$QE: $\alpha$ query expansion.  DFS: global diffusion. 
The best results are in \textbf{bold}.}
\label{tab:Rox5kpar6kSOTA}
\setlength{\tabcolsep}{1mm}
\resizebox{\textwidth}{!}{%
\begin{tabular}{lcccccccc}
\toprule
 & \multicolumn{4}{c}{Medium} & \multicolumn{4}{c}{Hard} \\
 \cmidrule(r){2-5} \cmidrule(r){6-9}
 & \multicolumn{2}{c}{R-Oxford} & \multicolumn{2}{c}{R-Paris} & \multicolumn{2}{c}{R-Oxford} & \multicolumn{2}{c}{R-Paris} \\
\cmidrule(r){2-3} \cmidrule(r){4-5} \cmidrule(r){6-7} \cmidrule(r){8-9}
Method & mAP & mP@10 & mAP & mP@10 & mAP & mP@10 & mAP & mP@10 \\
\cmidrule(r){1-9}
ResNet101--GeM+$\alpha$QE~\cite{Radenovic-arXiv17} & 67.2 & 86.0 & 80.7 & \textbf{98.9} & 40.7 & 54.9 & 61.8 & 90.6 \\
ResNet101--GeM\cite{Radenovic-arXiv17}+DFS~\cite{Iscen2017CVPR} & 69.8 & 84.0 & 88.9 & 96.9 & 40.5 & 54.4 & 78.5 & \textbf{94.6} \\
ResNet101--R-MAC\cite{Gordo-IJCV17}+DFS~\cite{Iscen2017CVPR} & 69.0 & 82.3 & \textbf{89.5} & 96.7 & 44.7 & 60.5 & \textbf{80.0} & 94.1 \\
ResNet50--DELF\cite{DELF2017}--HQE+SV & 73.4 & 88.2 & 84.0 & 98.3 & 50.3 & 67.2 & 69.3 & 93.7 \\
\cmidrule(r){1-9}
HesAff--RootSIFT--HQE~\cite{Tolias-PR2014HQE} & 66.3 & 85.6 & 68.9 & 97.3 & 41.3 & 60.0 & 44.7 & 79.9 \\
HesAff--RootSIFT--HQE+SV~\cite{Tolias-PR2014HQE} & 71.3 & 88.1 & 70.2 & 98.6 & 49.7 & 69.6 & 45.1 & 83.9 \\
\cmidrule(r){1-9}
HesAffNet--HardNet++--HQE & 71.7 & 89.4 & 72.6 & 98.1 & 47.5 & 66.3 & 48.9 & 85.9 \\
HesAffNet--HardNet++--HQE+SV & \textbf{75.2} & \textbf{90.9} & 73.1 & 98.1 & \textbf{53.3} &
\textbf{72.6} & 48.9 & 89.1 \\
\bottomrule
\end{tabular}
}
\end{table*}
We evaluate the proposed approach on standard image retrieval datasets Oxford5k~\cite{Philbin07} and Paris6k~\cite{Philbin08}.
Each dataset contains images (5062 for Oxford5k and 6391 for Paris6k) depicting 11 different landmarks and distractors. The performance is reported as mean average precision (mAP)~\cite{Philbin07}.
Recently, these benchmarks have been revisited, annotation errors fixed and new, more challenging sets of queries added~\cite{revisitop}. The revisited datasets define new test protocols: \textit{Easy}, \textit{Medium}, and \textit{Hard}.

We use the multi-scale Hessian-affine detector~\cite{Mikolajczyk2005} with the Baumberg method for affine shape estimation. The proposed AffNet replaces Baumberg, which we denote HessAffNet. The use of HessAffNet increased the number of used feature, from 12.5M to 17.5M for Oxford5k and from 15.6M to 21.2M for Paris6k, because more features survive the affine shape adaptions, as explained in Section~\ref{exp:rep}. We also performed additional experiment by restricting number of AffNet features to same as in Baumberg -- HesAffNetLess in Table~\ref{tab:ox5kpar6k}. We evaluated HesAffNet with both hand-crafted descriptor RootSIFT~\cite{RootSIFT2012} and state-of-the-art learned descriptors~\cite{TFeat2016,HardNet2017}.

First, HesAffNet is tested within the traditional bag-of-words~(BoW)~\cite{Sivic-ICCV2003VideoGoogle} image retrieval pipeline. A flat vocabulary with 1M centroids is created with the k-means algorithm and approximate nearest neighbor search~\cite{flann2009}. All descriptors of an image are assigned to a respective centroid of the vocabulary, and then they are aggregated with a histogram of occurrences into a BoW image representation.

We also apply spatial verification~(SV)~\cite{Philbin07} and standard query expansion~(QE)~\cite{Philbin08}. QE is performed with images that have either 15 (typically used) or 8 inliers after the spatial verification. 
 The results of the comparison are presented in Table~\ref{tab:ox5kpar6k}.

AffNet achieves the best results on both Oxford5k and Paris6k datasets, in most of the cases it outperforms the second best approach by a large margin. This experiment clearly shows the benefit of using AffNet in the local feature detection pipeline.

Additionally, we compare with state-of-the-art local-feature-based image retrieval methods.
A visual vocabulary of 65k words is learned, with Hamming embedding (HE)~\cite{Jegou-IJCV2010HE} technique added that further refines descriptor assignments with a 128 bits binary signature. We follow the same procedure as HesAff--RootSIFT--HQE~\cite{Tolias-PR2014HQE} method.
All parameters are set as in~\cite{Tolias-PR2014HQE}. 
The performance of AffNet methods is the best reported on both Oxford5k and Paris6k for local features. 

Finally, on the revisited R-Oxford and R-Paris, we compare with state-of-the-art methods in image retrieval, both local and global feature based: the best-performing fine-tuned networks~\cite{He2016ResNet}, ResNet101 with generalized-mean pooling (ResNet101--GeM)~\cite{Radenovic-arXiv17} and ResNet101 with regional maximum activations pooling (ResNet101--R-MAC)~\cite{Gordo-IJCV17}. Deep methods use re-ranking methods: $\alpha$ query expansion ($\alpha$QE)~\cite{Radenovic-arXiv17}, and global diffusion (DFS)~\cite{Iscen2017CVPR}. Results are in Table~\ref{tab:Rox5kpar6kSOTA}. 

HesAffNet performs best on the R-Oxford. It is consistently the best performing local-feature method, yet is worse than deep methods on R-Paris. A  possible explanation is that deep networks (ResNet and DELF) were finetuned from ImageNet, which contains Paris-related images, e.g. Sacre-Coeur and Notre Dame Basilica in the ``church'' category. Therefore global deep nets are partially evaluated on the training set. 

\section{Conclusions}
\label{sec:conclusions}
We presented a method for learning affine shape of local features in a weakly-supervised manner. The proposed HardNegC loss function might find other application domains as well. Our intuition is that the distance to the hard-negative estimates the local density of all points and provides a scale for the positive distance. 
The resulting AffNet regressor bridges the gap between performance of the similarity-covariant and affine-covariant detectors on images with short baseline and big illumination differences and it improves performance of affine-covariant detectors in the wide baseline setup. AffNet applied to the output of the Hessian detector improves the state-of-the art in wide baseline matching, affine detector repeatability and image retrieval.

We experimentally show that descriptor matchability, not only repeatability should be taken into account when learning a feature detector.

\textbf{Acknowledgements}
The authors were supported by the Czech Science Foundation Project GACR P103/12/G084, the Austrian Ministry for Transport, Innovation and Technology, the Federal Ministry of Science, Research and Economy, and the Province of Upper Austria in the frame of the COMET center SCCH, the CTU student grant SGS17/185/OHK3/3T/13, and the MSMT LL1303 ERC-CZ grant. 
\clearpage
{\small
\bibliographystyle{splncs}

\input{eccv2018submission.bbl}
}
\end{document}

%% file: includes.tex
\usepackage{graphicx}
\usepackage{amsmath,amssymb} 
\usepackage{color}
\usepackage{lscape}
\usepackage{xspace}
\usepackage{times}
\usepackage{epsfig}
\usepackage{mathtools}
\usepackage{float}
\usepackage{graphicx}
\usepackage{amsmath}
\usepackage[table]{xcolor}
\usepackage{amssymb}
\usepackage{booktabs}
\usepackage[utf8]{inputenc} 
\usepackage[T1]{fontenc}  
\usepackage{url}      
\usepackage{booktabs}    
\usepackage{amsfonts}    
\usepackage{nicefrac}    
\usepackage{microtype}   
\usepackage{amsmath}

\DeclareMathOperator*{\argmin}{arg\,min}
\usepackage[width=122mm,left=12mm,paperwidth=146mm,height=193mm,top=12mm,paperheight=217mm]{geometry}
\usepackage[pagebackref=true,breaklinks=true,colorlinks,bookmarks=false]{hyperref}
\makeatletter
\DeclareRobustCommand\onedot{\futurelet\@let@token\@onedot}
\def\@onedot{\ifx\@let@token.\else.\null\fi\xspace}
\newcommand{\ra}[1]{\renewcommand{\arraystretch}{#1}}
\def\eg{\emph{e.g}\onedot} 
\def\ie{\emph{i.e}\onedot}

\def\etal{\emph{et al}\onedot}

\makeatother

%% file: intro.tex
\section{Introduction}
\label{sec:intro}

Local features, forming correspondences, are exploited in state of the art pipelines 
for 3D reconstruction~\cite{Schonberger-CVPR2016sfm,schoenberger2017}, two-view matching~\cite{MODS2015}, 6DOF image localization~\cite{6DOFPlace2017}. Classical local features have also been successfully used for providing supervision for CNN-based image retrieval~\cite{Radenovic-ECCV2016CNNfromBOW}. 

Affine-convariance~\cite{Mikolajczyk2004} is a desirable property of local features
since it allows robust matching of images separated by a wide baseline~\cite{Mikolajczyk2005,MODS2015}, unlike scale-covariant features like ORB~\cite{Rublee2011} or difference of Gaussian (DoG)~\cite{SIFT2004} that rely on tests carried out on circular neighborhoods. This is the reason why the Hessian-Affine detector~\cite{Mikolajczyk2004} combined with the RootSIFT descriptor~\cite{SIFT2004,RootSIFT2012} is the gold standard for local feature in image retrieval~\cite{Perdoch-CVPR2009efficient,Tolias-PR2014HQE}. Affine covariant features also provide stronger geometric constraints, e.g., for image rectification~\cite{RepPat2018}.

On the other hand, the classical affine adaptation procedure~\cite{Baumberg2000} fails in 20\%-40\%~\cite{Mikolajczyk2005,WXBS2015} cases, thus reducing the number and repeatability of detected local features.
It is also not robust to significant illumination change~\cite{WXBS2015}. Applications where the number of detected features is important,~\eg, large scale 3D reconstruction~\cite{schoenberger2017}, therefore use the DoG detector. 
Alleviating the problem of the drop in the number of correspondences caused by the non-repeatability of the affine adaptation procedure,
may lead to connected 3D reconstructions and improved image retrieval engines~\cite{Schonberger-CVPR2015retrievalsfm,radenovic2016dusk}.

This paper makes four contributions towards robust estimation of the local affine shape. 
First, we experimentally show that geometric repeatability of a local feature is not a sufficient condition for successful matching. The learning of affine shape increases the number of corrected matches if it steers the estimators towards discriminative regions and therefore must involve optimization of a descriptor-related loss.

Second, we propose a novel loss function for descriptor-based registration and learning, named the {\it hard negative-constant loss}. It combines the advantages of the triplet and contrastive positive losses. Third, we propose a method for learning the affine shape, orientation and potentially other parameters related to  geometric and appearance properties of local features. The learning method  does not require a precise ground truth which reduces the need for manual annotation. 

Last but not least, the learned AffNet itself significantly outperforms prior methods for affine shape estimation and improves the state of art in image retrieval by a large margin. Importantly, unlike the
de-facto standard~\cite{Baumberg2000}, AffNet does not significantly reduce the number of detected features, it is thus suitable even for pipelines where affine invariance is needed only occasionally.
\subsection{Related work}
The area of learning local features has been active recently, but the attention has focused dominantly on  learning descriptors~\cite{DeepComp2015,MatchNet2015,TFeat2016,L2Net2017,HardNet2017,SpreadDesc2017,ExemplarNet2016} and translation-covariant detectors~\cite{TIlde2015,NewTilde2017,Lenc2016,QuadNets2017}. The authors are not aware of any recent work on learning or improvement of local feature affine shape estimation. 
The most closely related work is thus the following.

Hartmann~\etal~\cite{HartmannHS14} train random forest classifier for predicting feature matchability based on a local descriptor. "Bad" points are discarded, thus speeding up the matching process in a 3D reconstruction  pipeline. 
Yi~\etal\cite{OriNet2016} proposed to learn feature orientation by minimizing descriptor distance between positive patches, i.e. those corresponding to the same point on the 3D surface. This allows to avoid hand-picking a "canonical" orientation, thus learning the one which is the most suitable for descriptor matching.  We have observed that direct application of the method~\cite{OriNet2016}  for affine shape estimation leads to learning degenerate shapes collapsed to single line. 
%
%
Yi~\etal\cite{LIFT2016} proposed a multi-stage framework for learning  the descriptor, orientation and translation-covariant detector. The detector was trained by maximizing the intersection-over-union and the reprojection error between corresponding regions.

Lenc and Vedaldi~\cite{Lenc2016} introduced the ``covariant constraint'' for learning various types of local feature detectors. The proposed covariant loss is the Frobenius norm of the difference between  the local affine frames. The disadvantage of such approach is that it could lead to features that are, while being repeatable, not necessarily suited for the matching task (see Section~\ref{sec:loss}).
On top of that, the common drawback of the Yi~\etal~\cite{LIFT2016} and Lenc and Vedaldi~\cite{Lenc2016} methods is that they require to know the exact geometric relationship between patches which increases the amount of work needed to prepare the training dataset. 
Zhang~\etal~\cite{NewTilde2017} proposed to ``anchor'' the detected features to some pre-defined features with known good discriminability like TILDE~\cite{TIlde2015}. We remark that despite showing images of affine-covariant features, the results presented in the paper are for translation-covariant features only. 
Savinov~\etal~\cite{QuadNets2017} proposed a ranking approach for unsupervised learning of a feature detector. While this is natural and efficient for learning the coordinates of the center of the feature, it is problematic to apply it for the affine shape estimation. The reason is that it requires sampling and scoring of many possible shapes.

Finally, Choy~\etal~\cite{UCN2016} trained a ``Universal correspondence network'' (UCN) for a direct correspondence estimation with contrastive loss on a patch descriptor distance. This approach is related to the current work, yet the two methods differ in several important aspects. First, UCN used an ImageNet-pretrained network which is subsequently fine-tuned. We learn the affine shape estimation from scratch. Second, UCN uses  dense feature extraction and negative examples extracted from the same image. While this could be a good setup for short baseline stereo, it does not work well for wide baseline, where affine features are usually sought. Finally, we propose the hard negative-constant loss instead of the contrastive one.  

%% file: fig-direct-opt-ideal-joint.tex
\begin{minipage}[h]{0.22\linewidth}
\center{\includegraphics[width=1\linewidth]{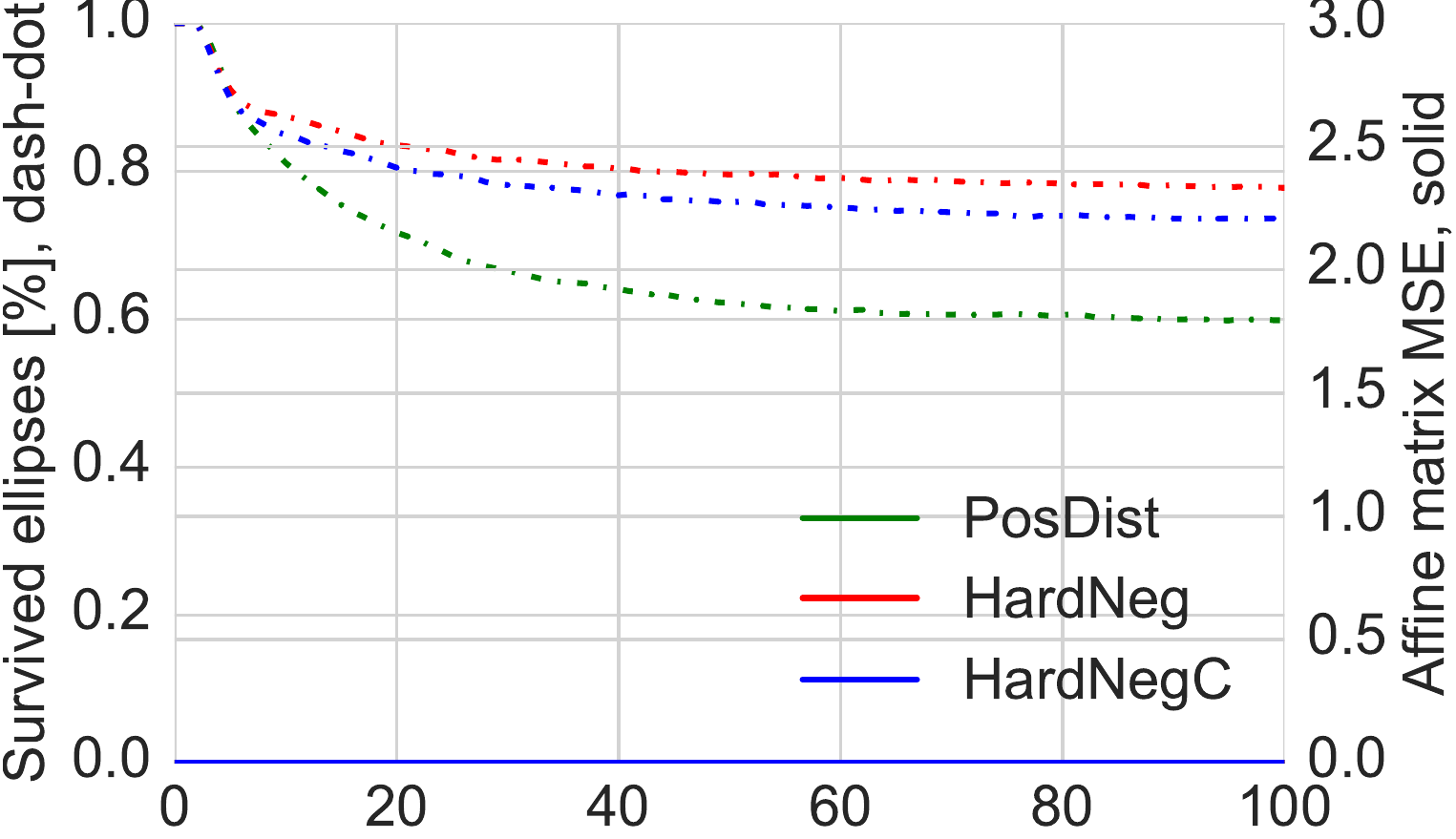}}  \\
\end{minipage}
\hfill
\begin{minipage}[h]{0.22\linewidth}
\center{\includegraphics[width=1\linewidth]{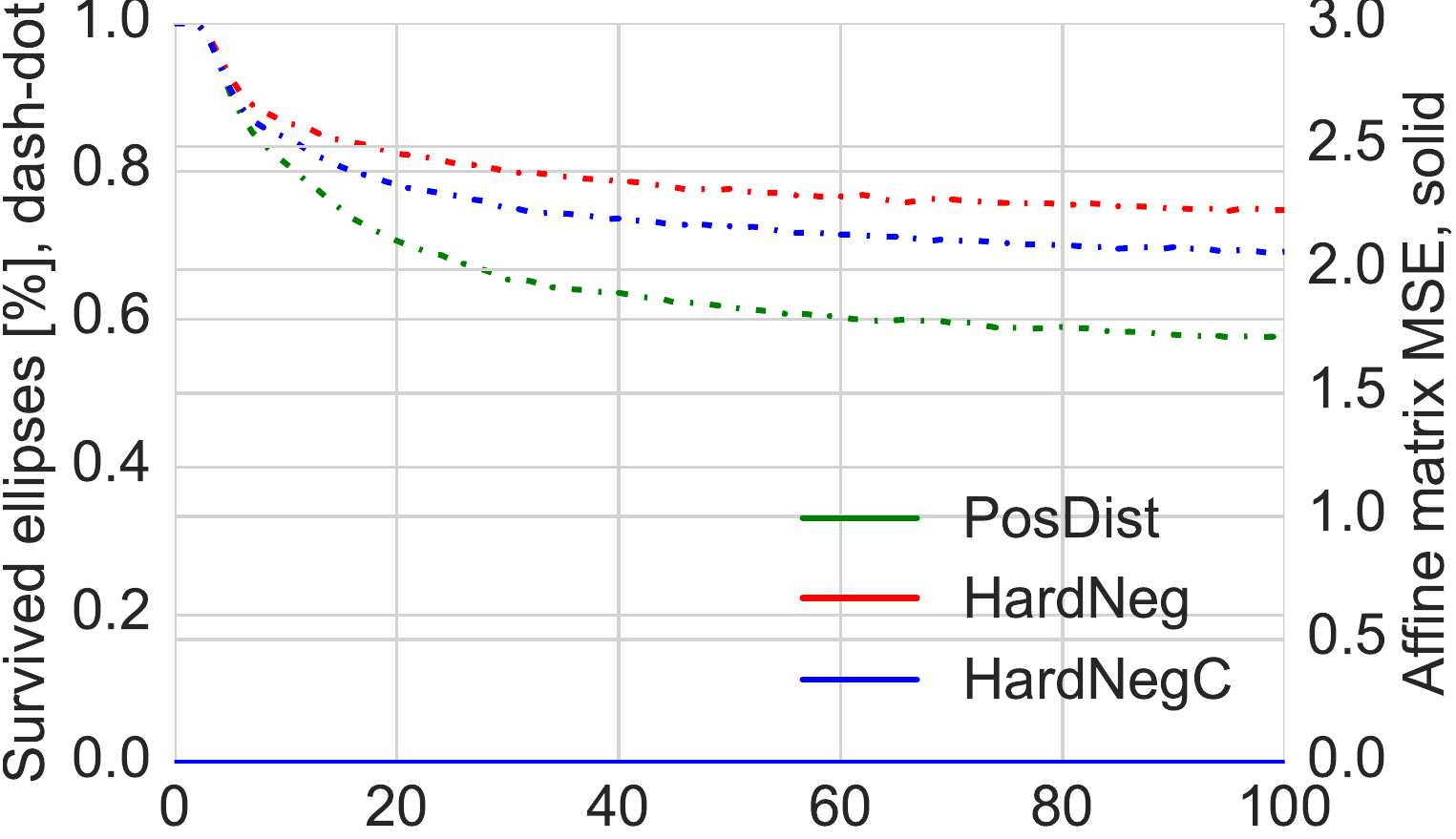}}  \\
\end{minipage}
\hfill
\begin{minipage}[h]{0.22\linewidth}
\center{\includegraphics[width=1\linewidth]{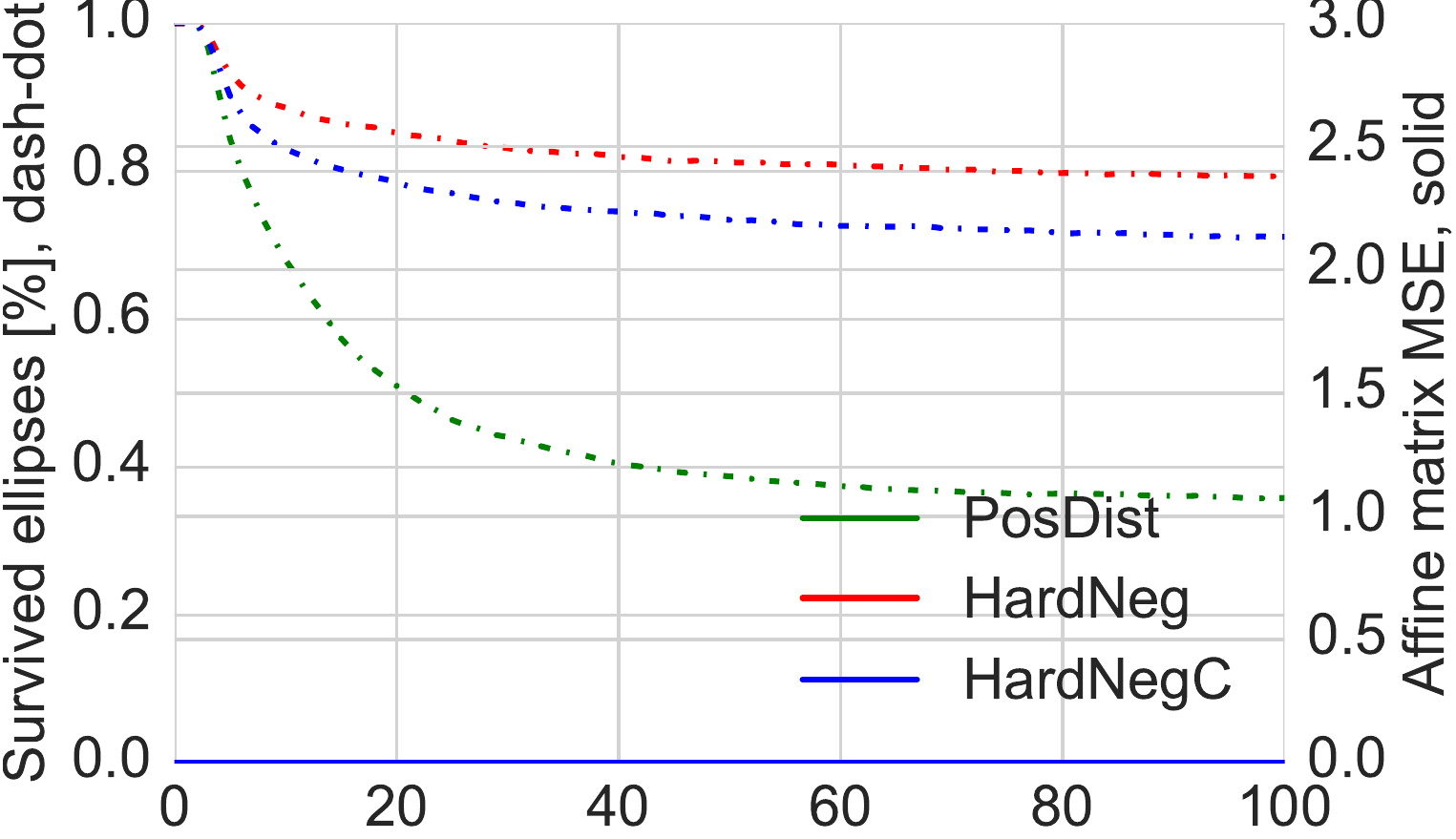}}  \\
\end{minipage}
\hfill
\begin{minipage}[h]{0.22\linewidth}
\center{\includegraphics[width=1\linewidth]{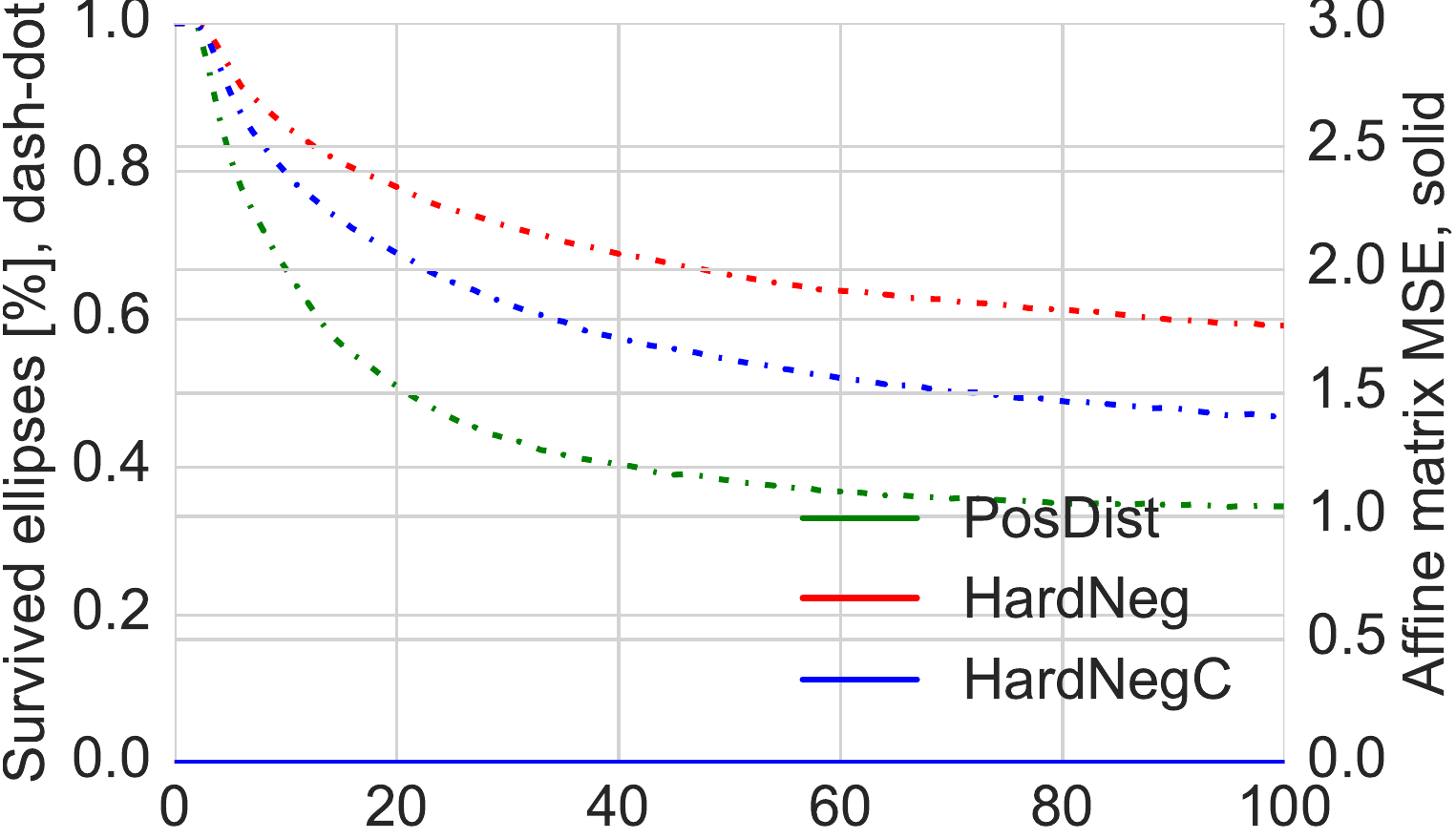}}  \\
\end{minipage}
\vfill
\begin{minipage}[h]{0.22\linewidth}
\center{\includegraphics[width=1\linewidth]{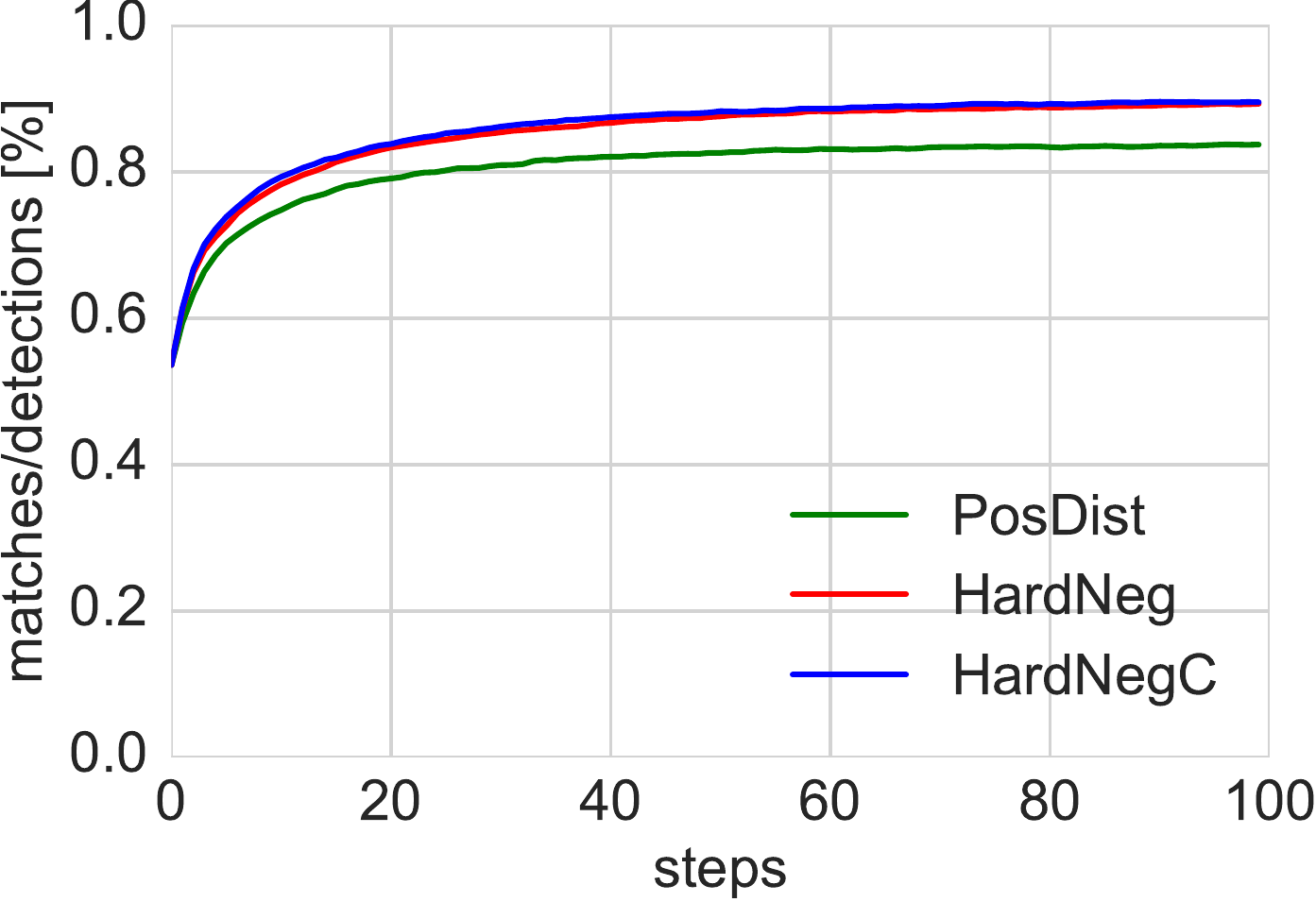}}  \\
\end{minipage}
\hfill
\begin{minipage}[h]{0.22\linewidth}
\center{\includegraphics[width=1\linewidth]{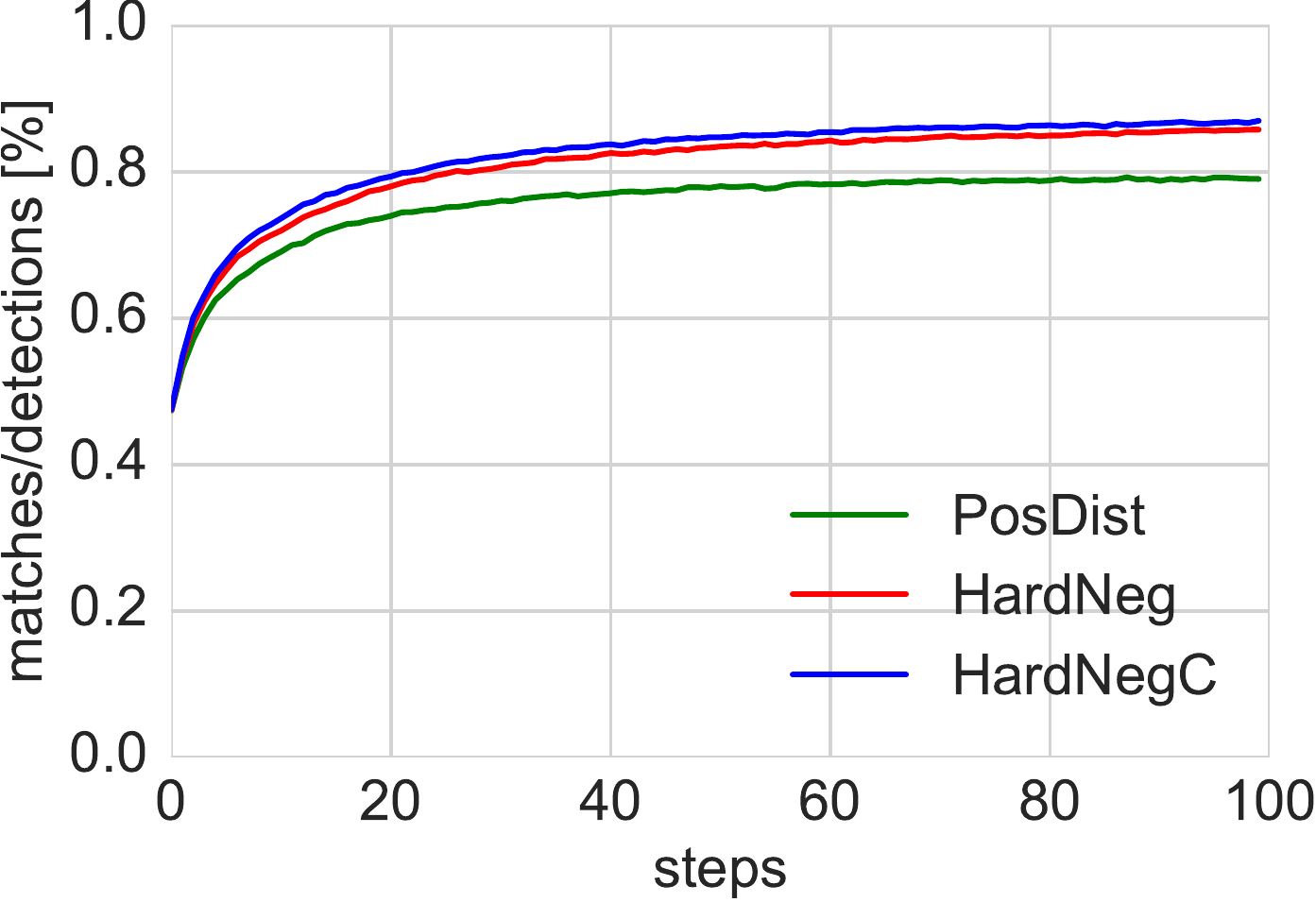}} \\
\end{minipage}
\hfill
\begin{minipage}[h]{0.22\linewidth}
\center{\includegraphics[width=1\linewidth]{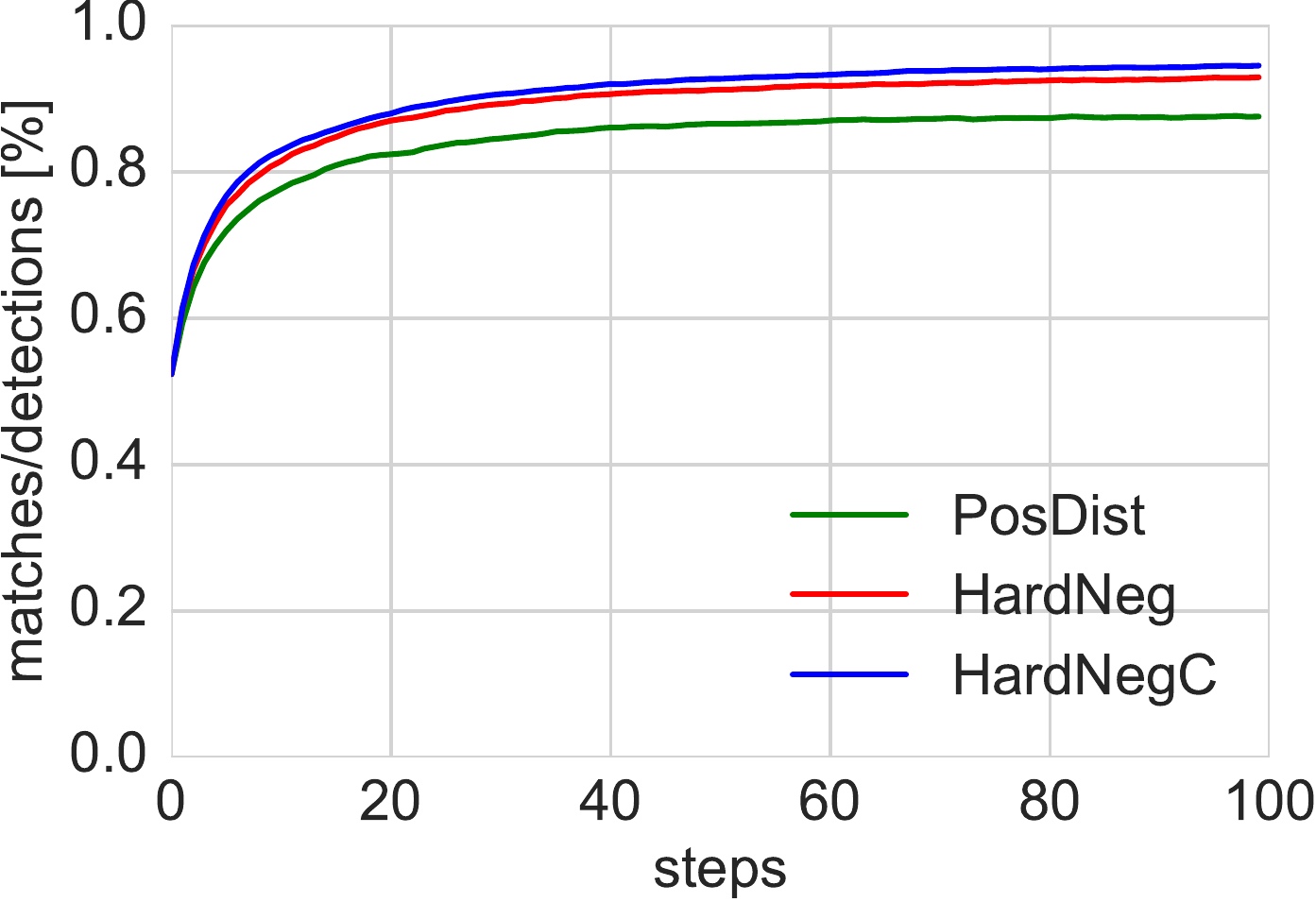}} \\
\end{minipage}
\hfill
\begin{minipage}[h]{0.22\linewidth}
\center{\includegraphics[width=1\linewidth]{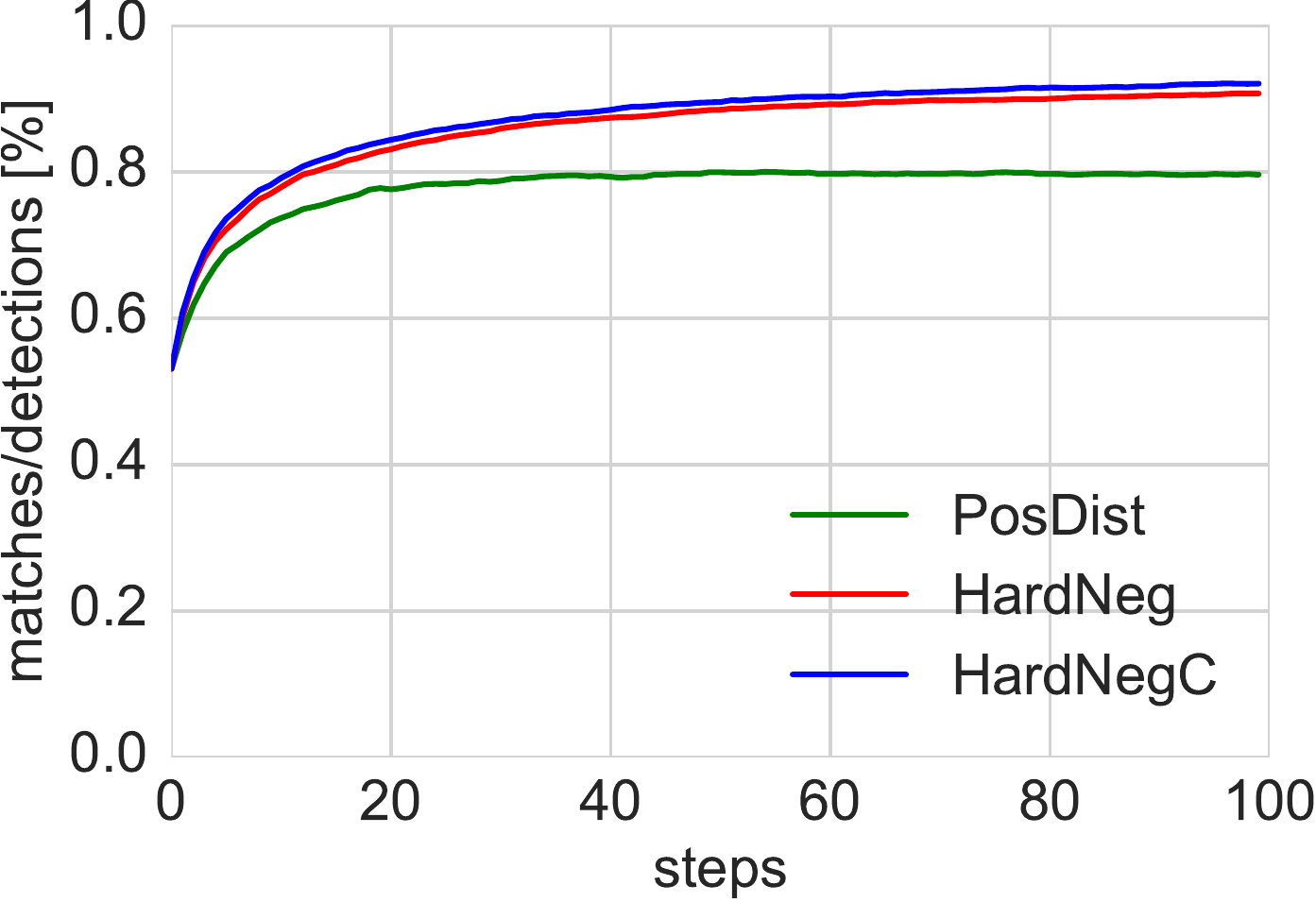}} \\
\end{minipage}

%% file: fig-direct-opt-ideal.tex
 \centering
 \begin{minipage}[h]{0.22\linewidth}
\center{\includegraphics[width=1\linewidth]{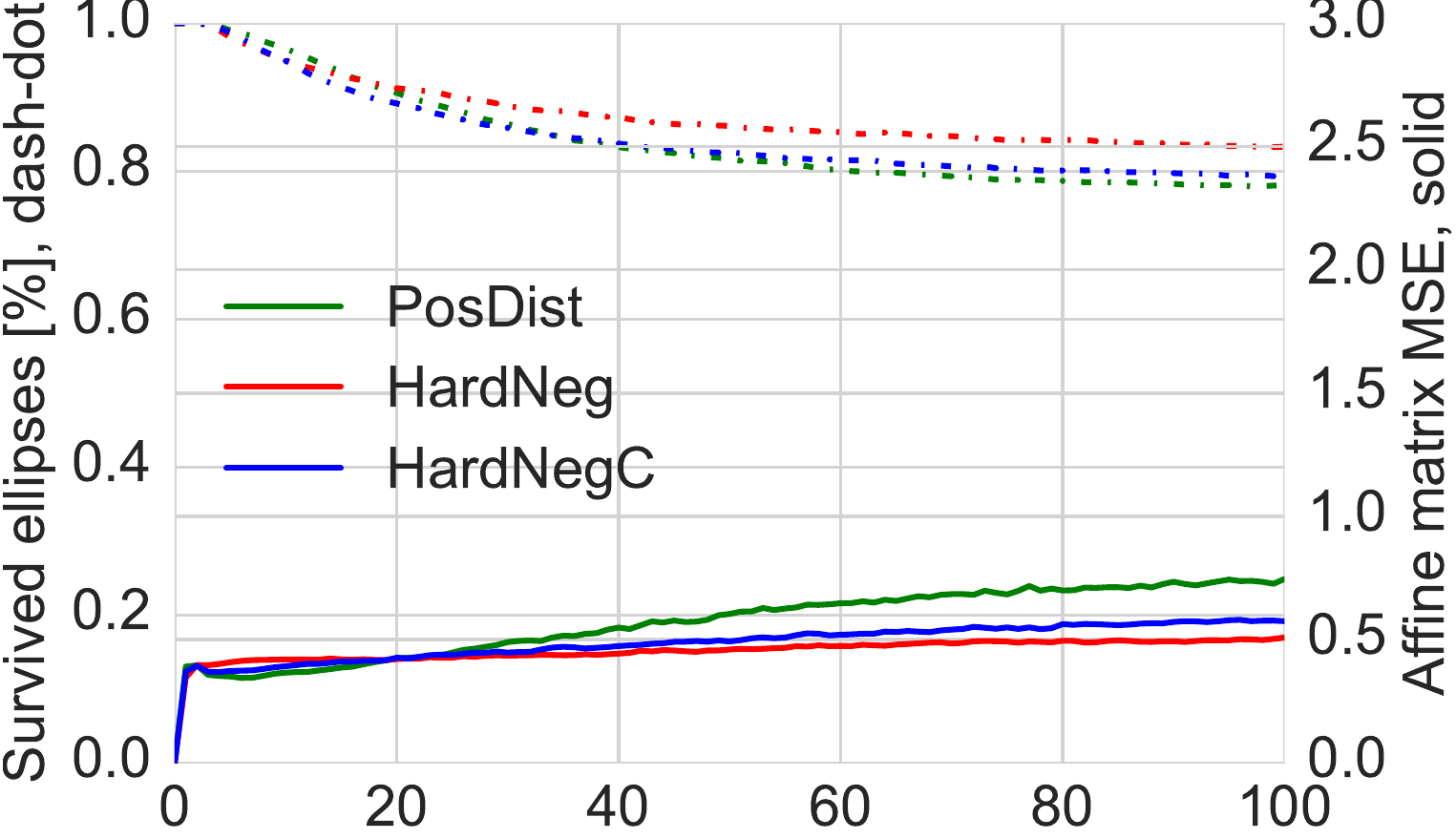}}  \\
\end{minipage}
\hfill
\begin{minipage}[h]{0.22\linewidth}
\center{\includegraphics[width=1\linewidth]{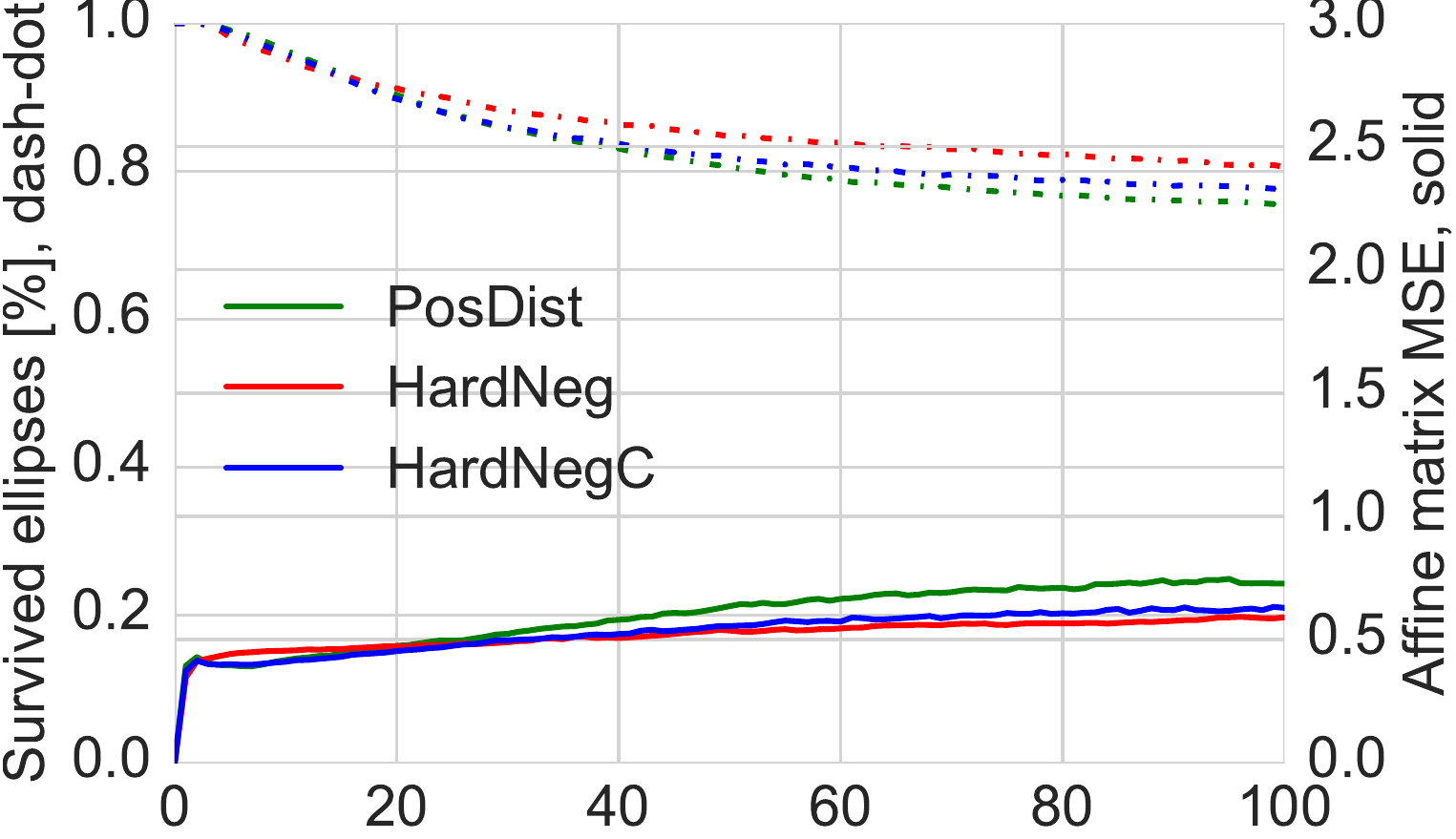}}  \\
\end{minipage}
\hfill
\begin{minipage}[h]{0.22\linewidth}
\center{\includegraphics[width=1\linewidth]{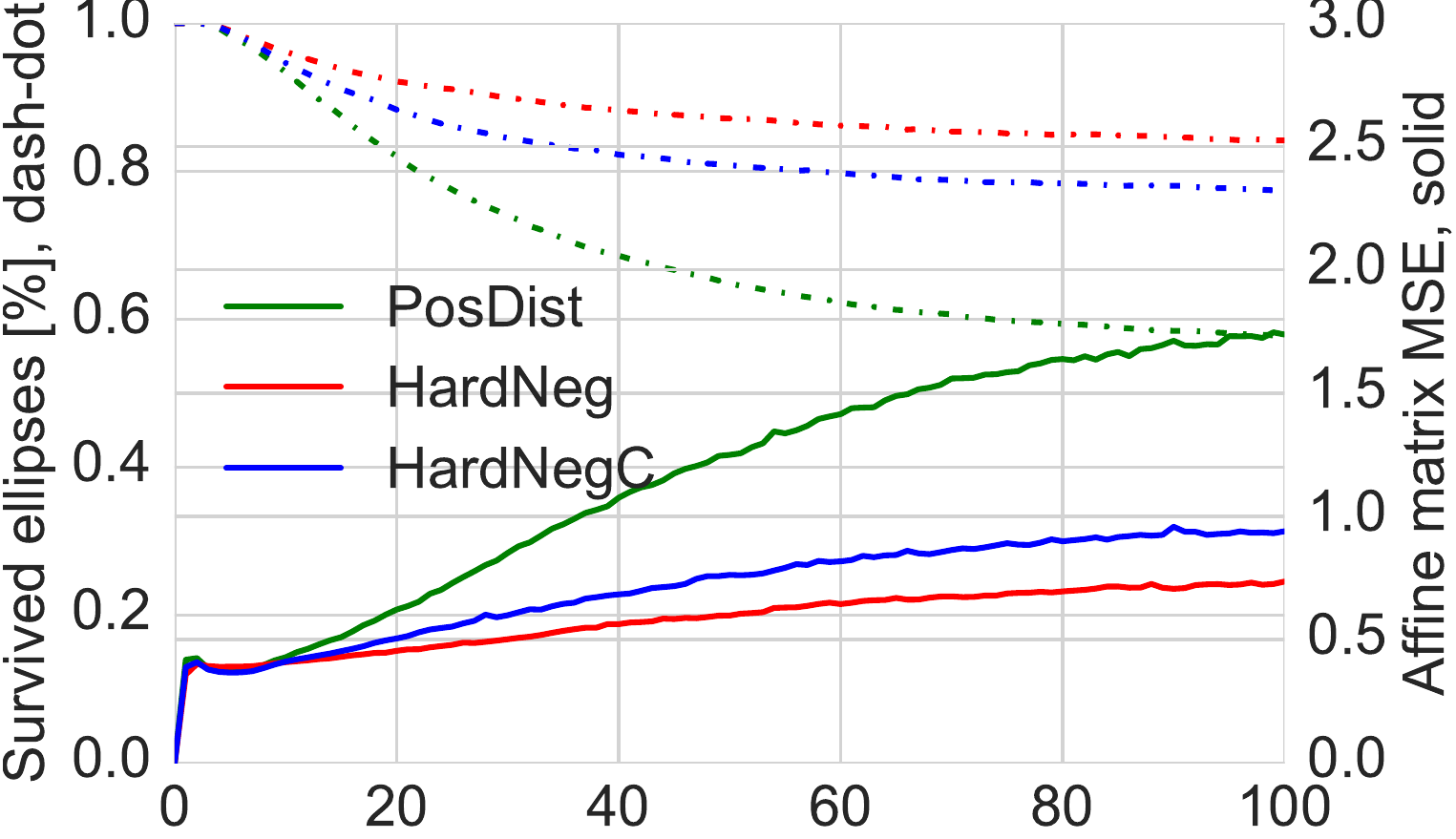}}  \\
\end{minipage}
\hfill
\begin{minipage}[h]{0.22\linewidth}
\center{\includegraphics[width=1\linewidth]{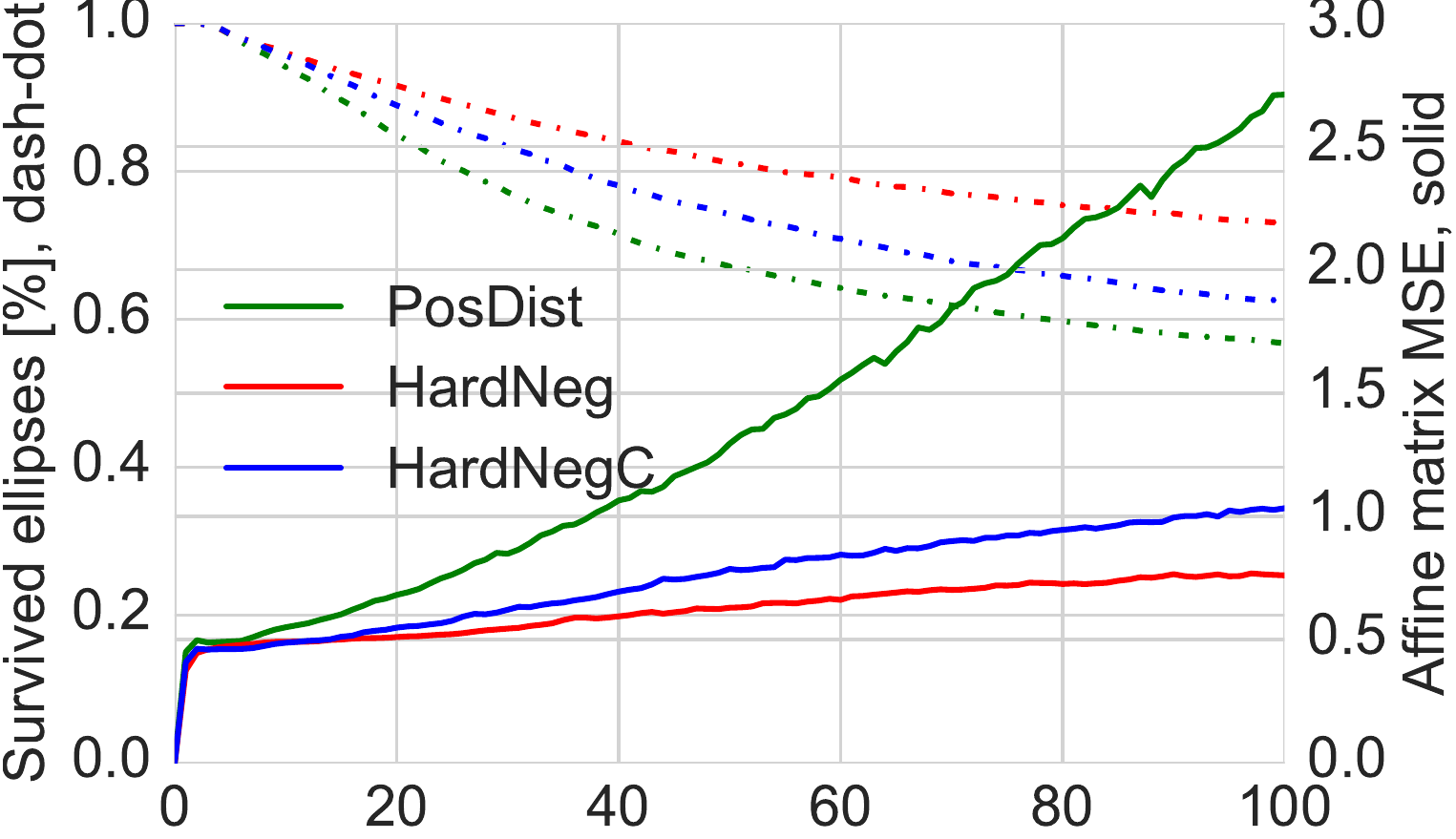}}  \\
\end{minipage}
\vfill
\begin{minipage}[h]{0.22\linewidth}
\center{\includegraphics[width=1\linewidth]{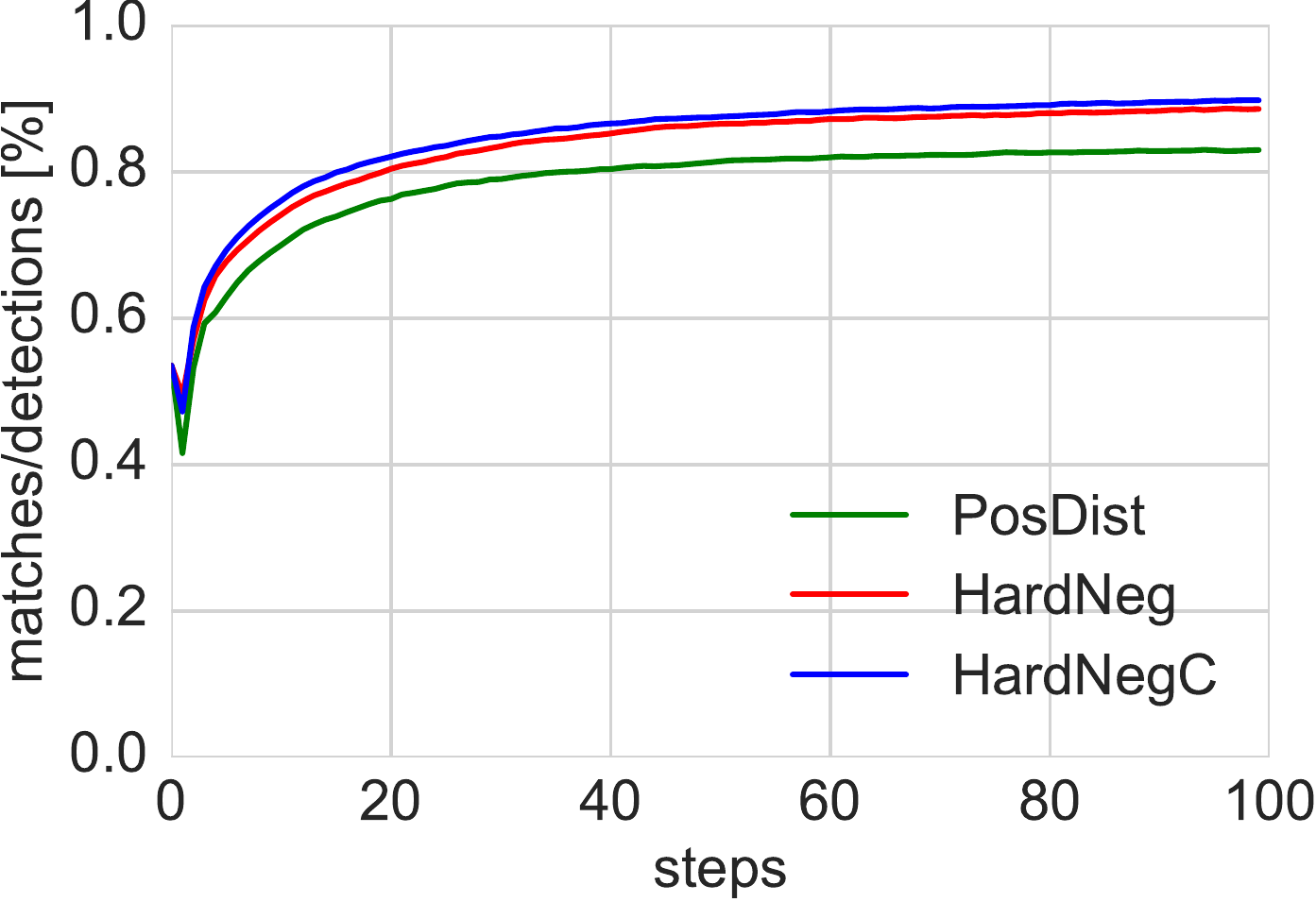}} HardNet \\
\end{minipage}
\hfill
\begin{minipage}[h]{0.22\linewidth}
\center{\includegraphics[width=1\linewidth]{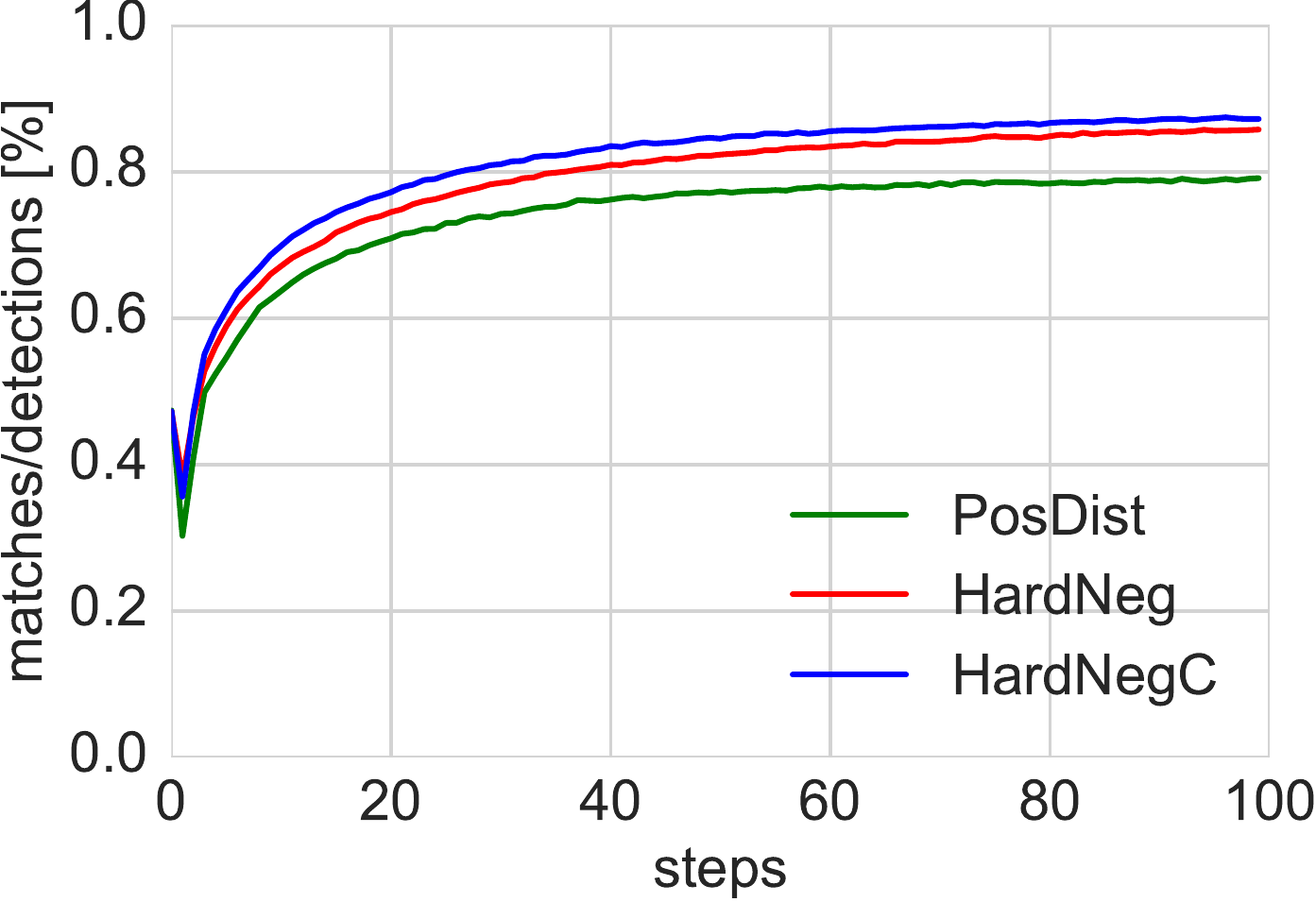}} TFeat \\
\end{minipage}
\hfill
\begin{minipage}[h]{0.22\linewidth}
\center{\includegraphics[width=1\linewidth]{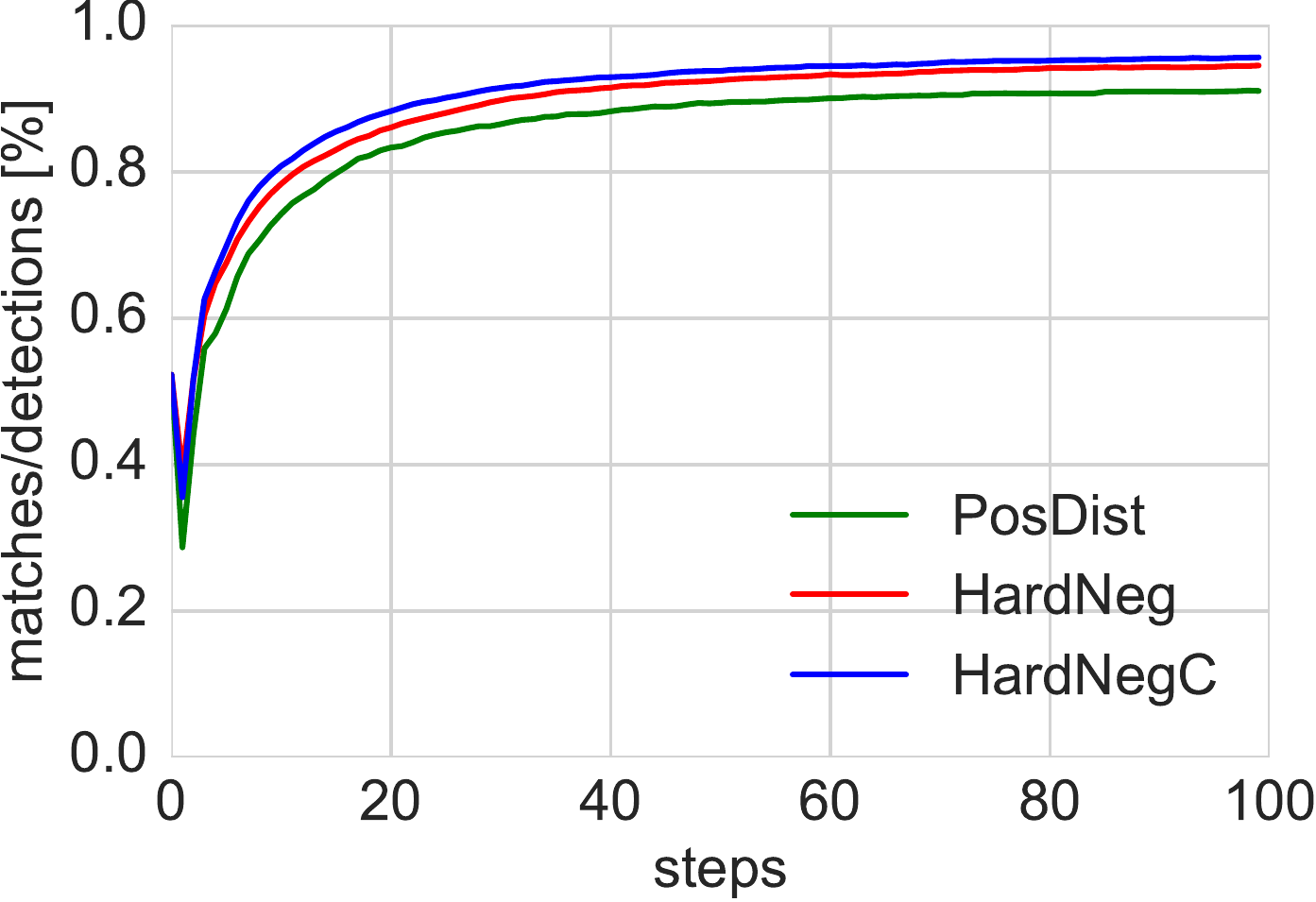}} SIFT \\
\end{minipage}
\hfill
\begin{minipage}[h]{0.22\linewidth}
\center{\includegraphics[width=1\linewidth]{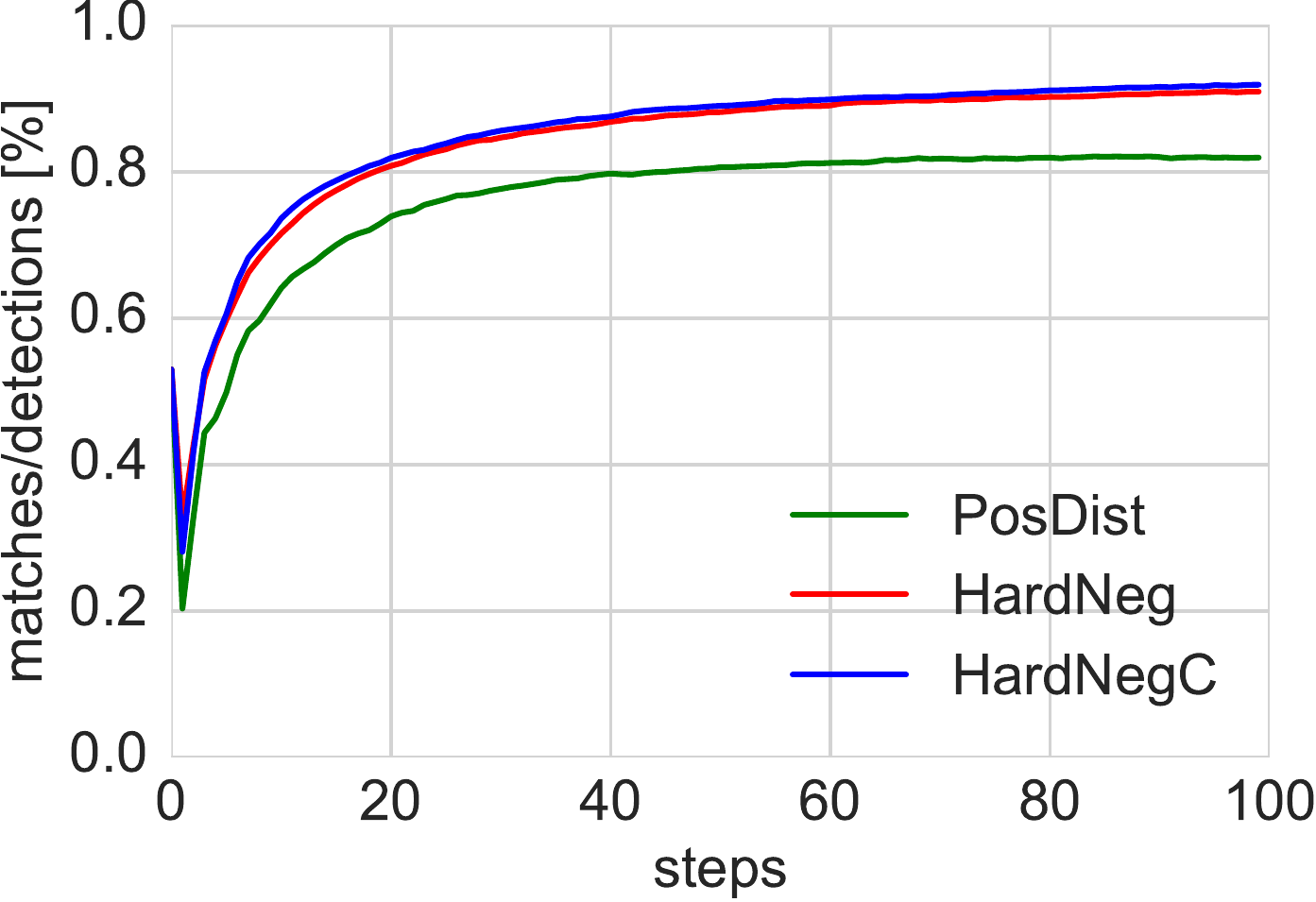}} Pixels  \\
\end{minipage}

%% file: fig-direct-opt-noise.tex
 \begin{minipage}[h]{0.24\linewidth}
\center{\includegraphics[width=1\linewidth]{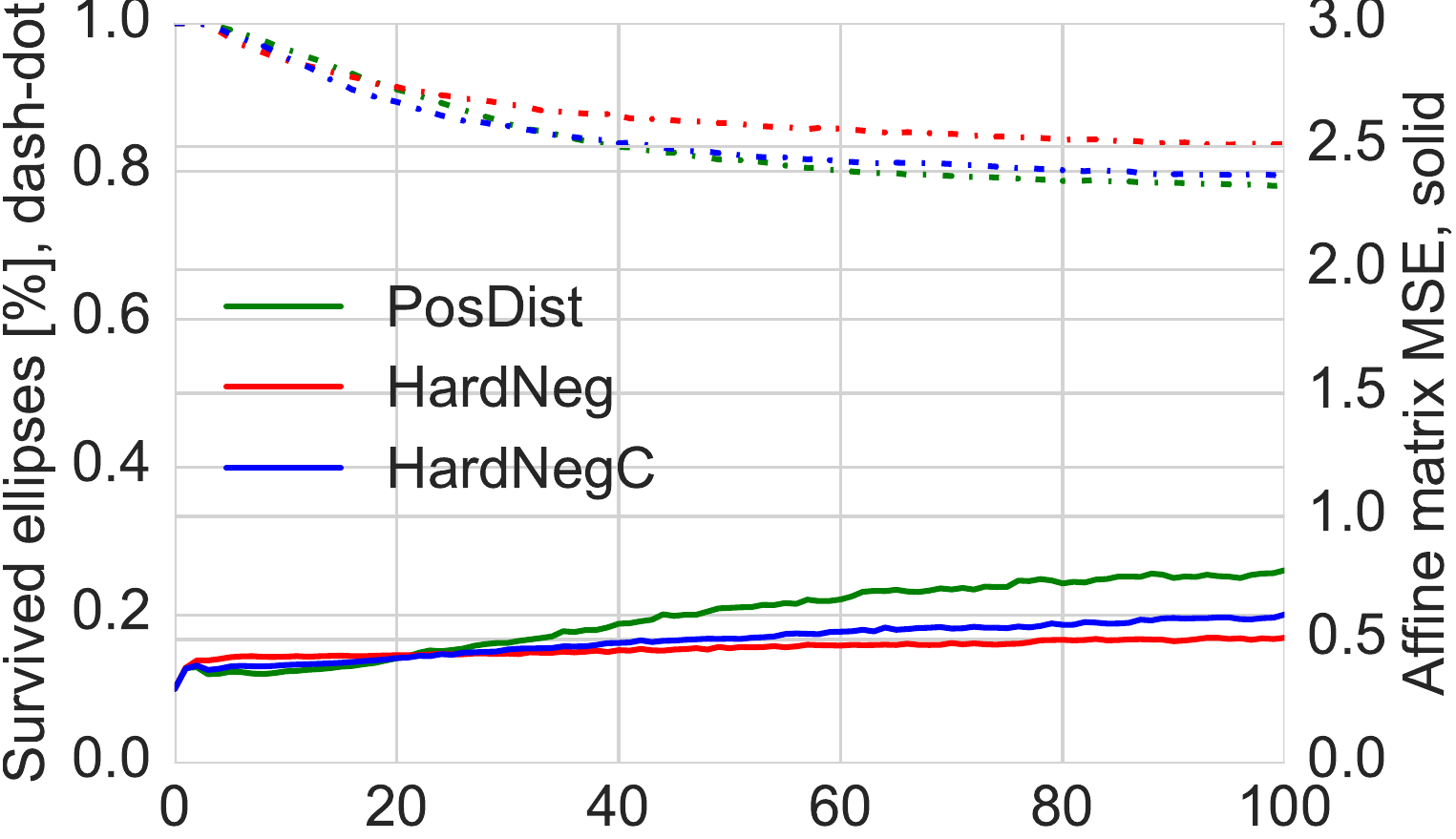}}  \\
\end{minipage}
\hfill
\begin{minipage}[h]{0.24\linewidth}
\center{\includegraphics[width=1\linewidth]{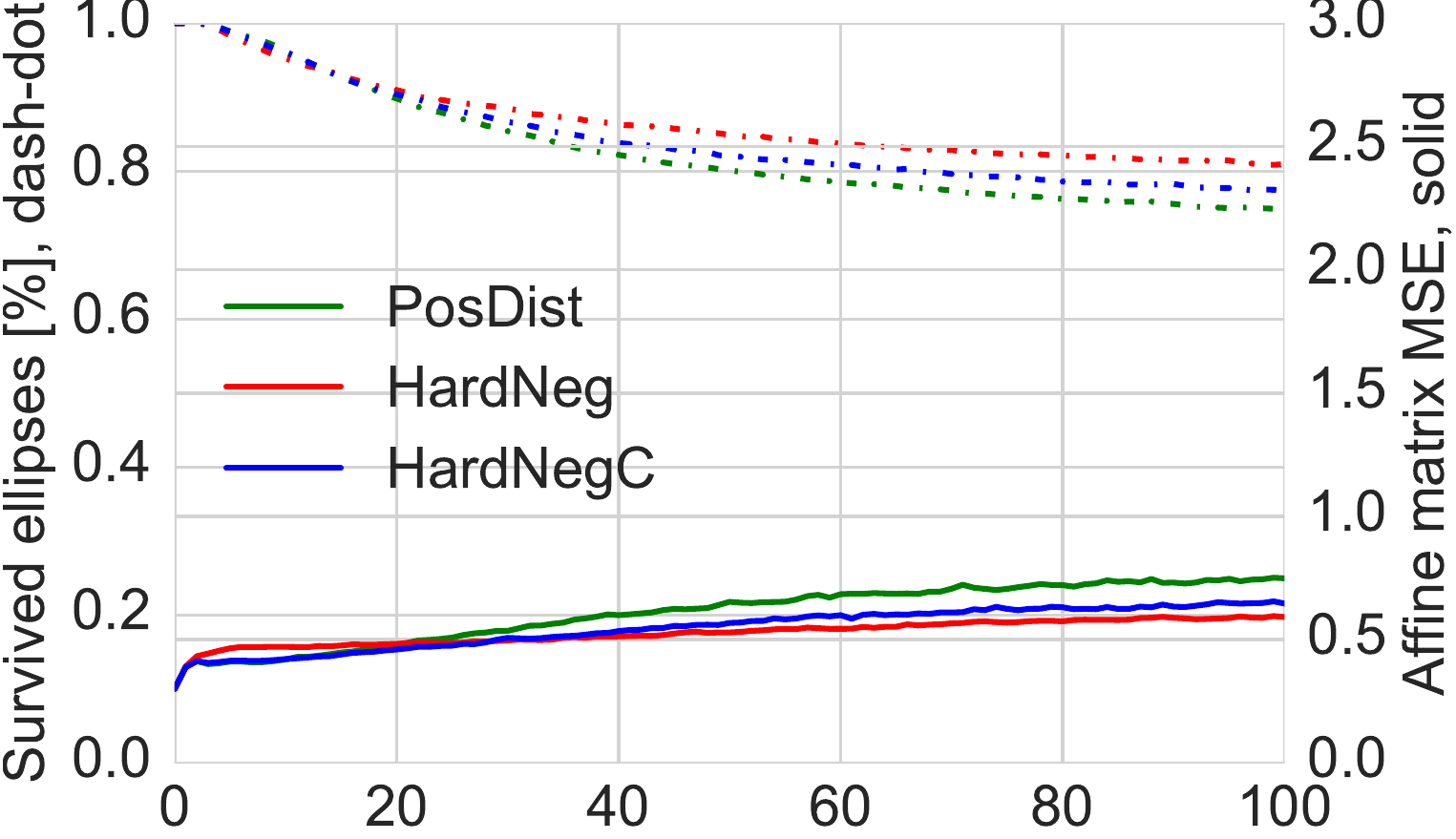}}  \\
\end{minipage}
\hfill
\begin{minipage}[h]{0.24\linewidth}
\center{\includegraphics[width=1\linewidth]{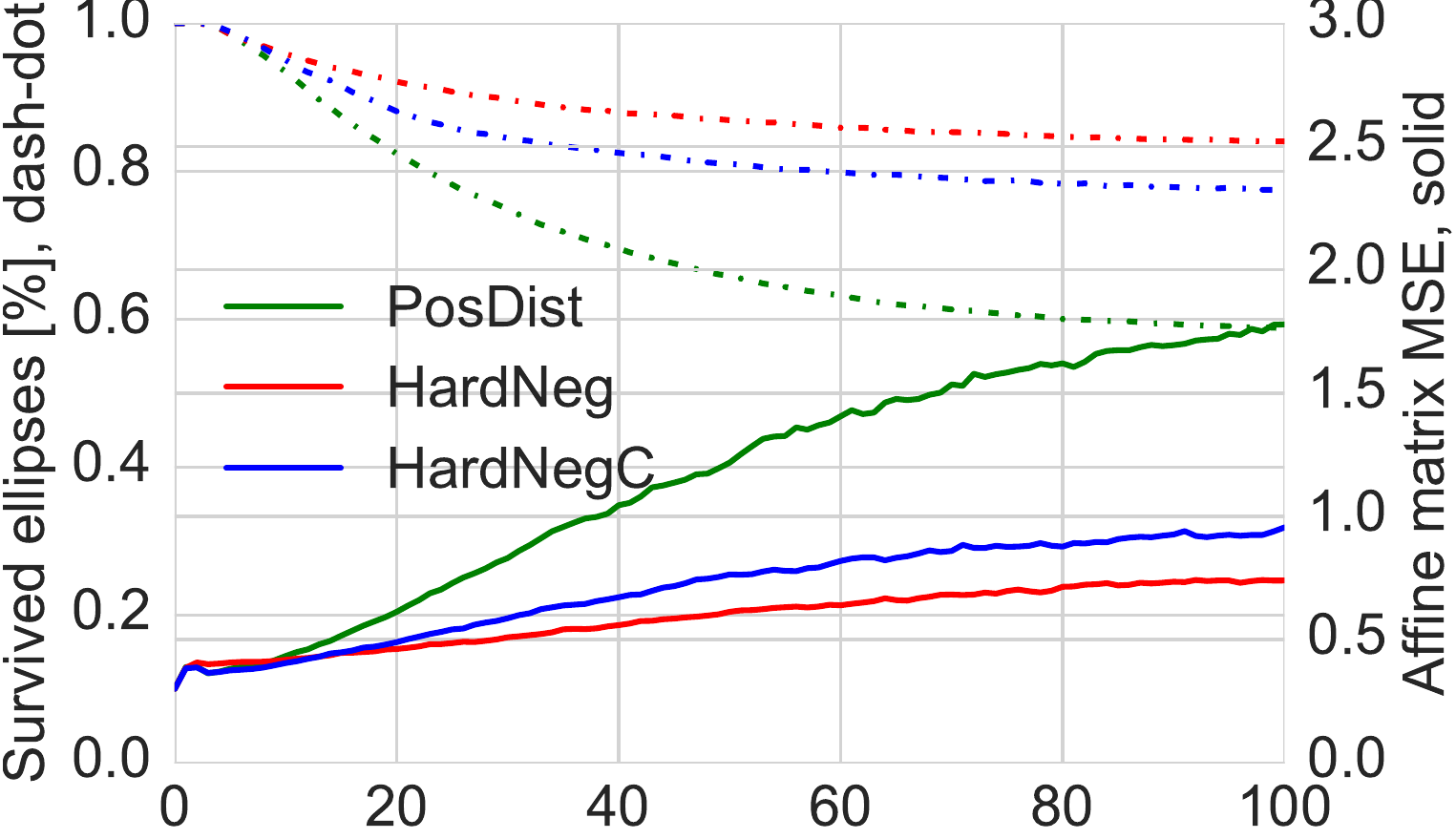}}  \\
\end{minipage}
\hfill
\begin{minipage}[h]{0.24\linewidth}
\center{\includegraphics[width=1\linewidth]{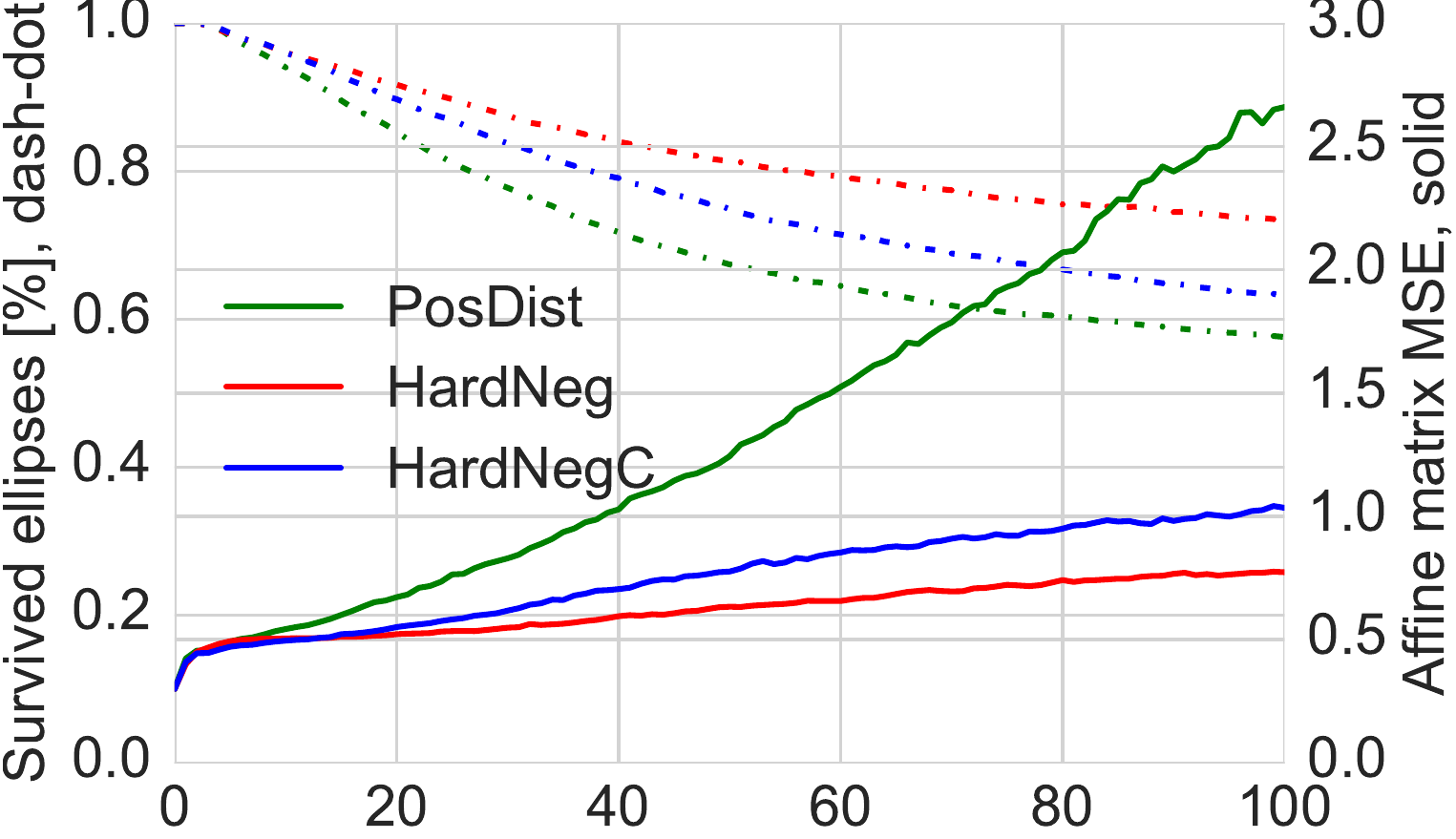}}  \\
\end{minipage}
\vfill
\begin{minipage}[h]{0.24\linewidth}
\center{\includegraphics[width=1\linewidth]{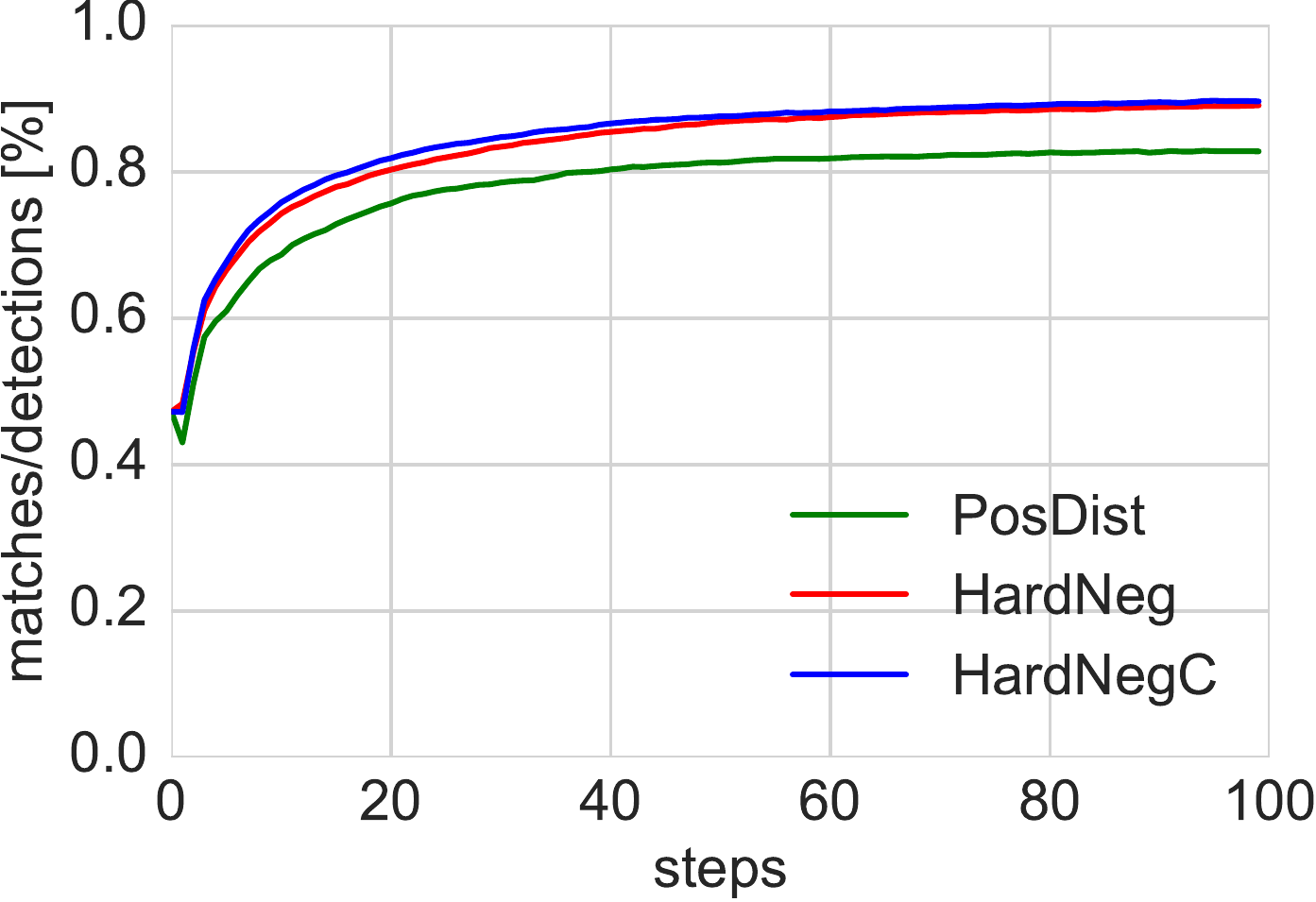}} HardNet \\
\end{minipage}
\hfill
\begin{minipage}[h]{0.24\linewidth}
\center{\includegraphics[width=1\linewidth]{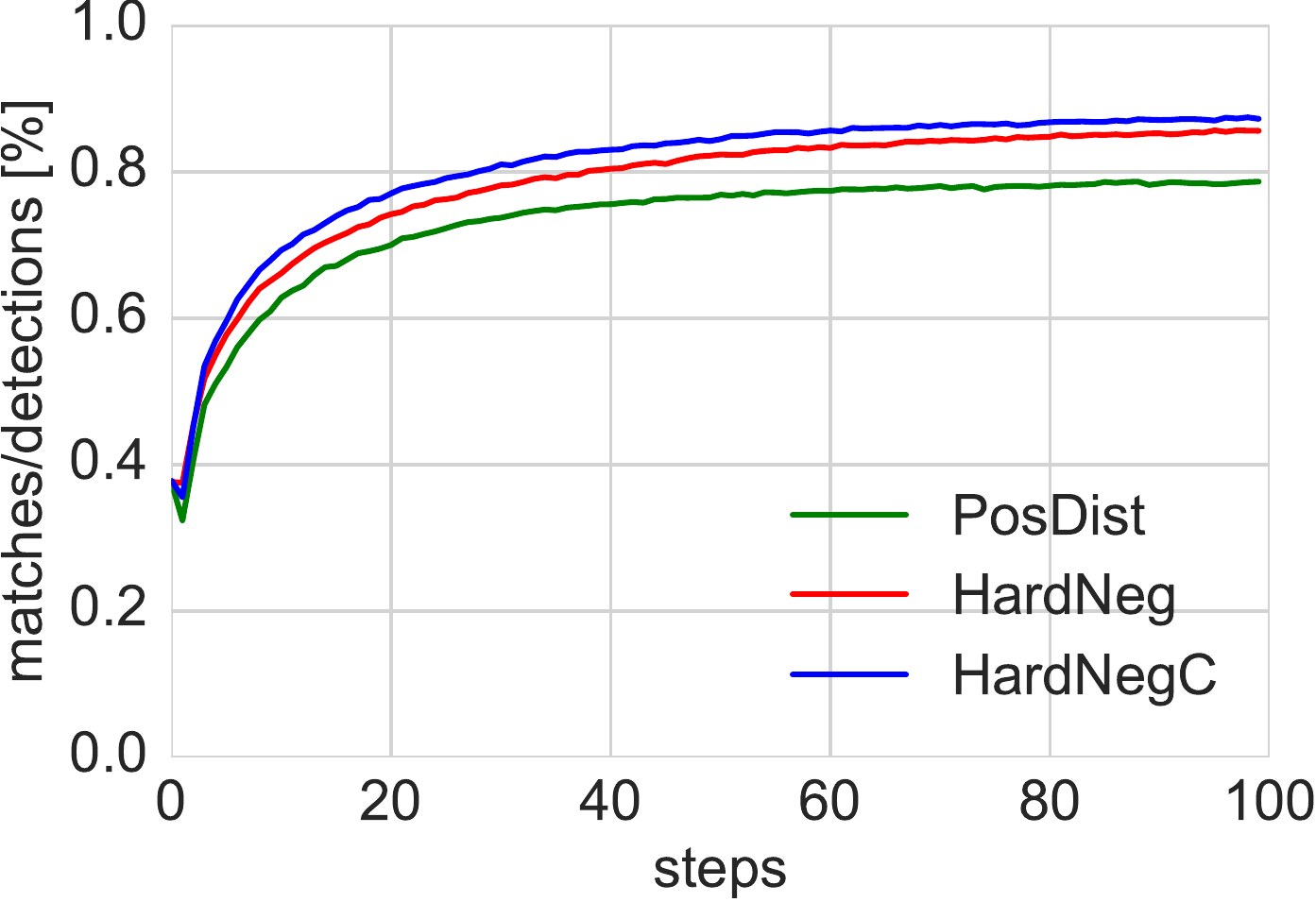}} TFeat \\
\end{minipage}
\hfill
\begin{minipage}[h]{0.24\linewidth}
\center{\includegraphics[width=1\linewidth]{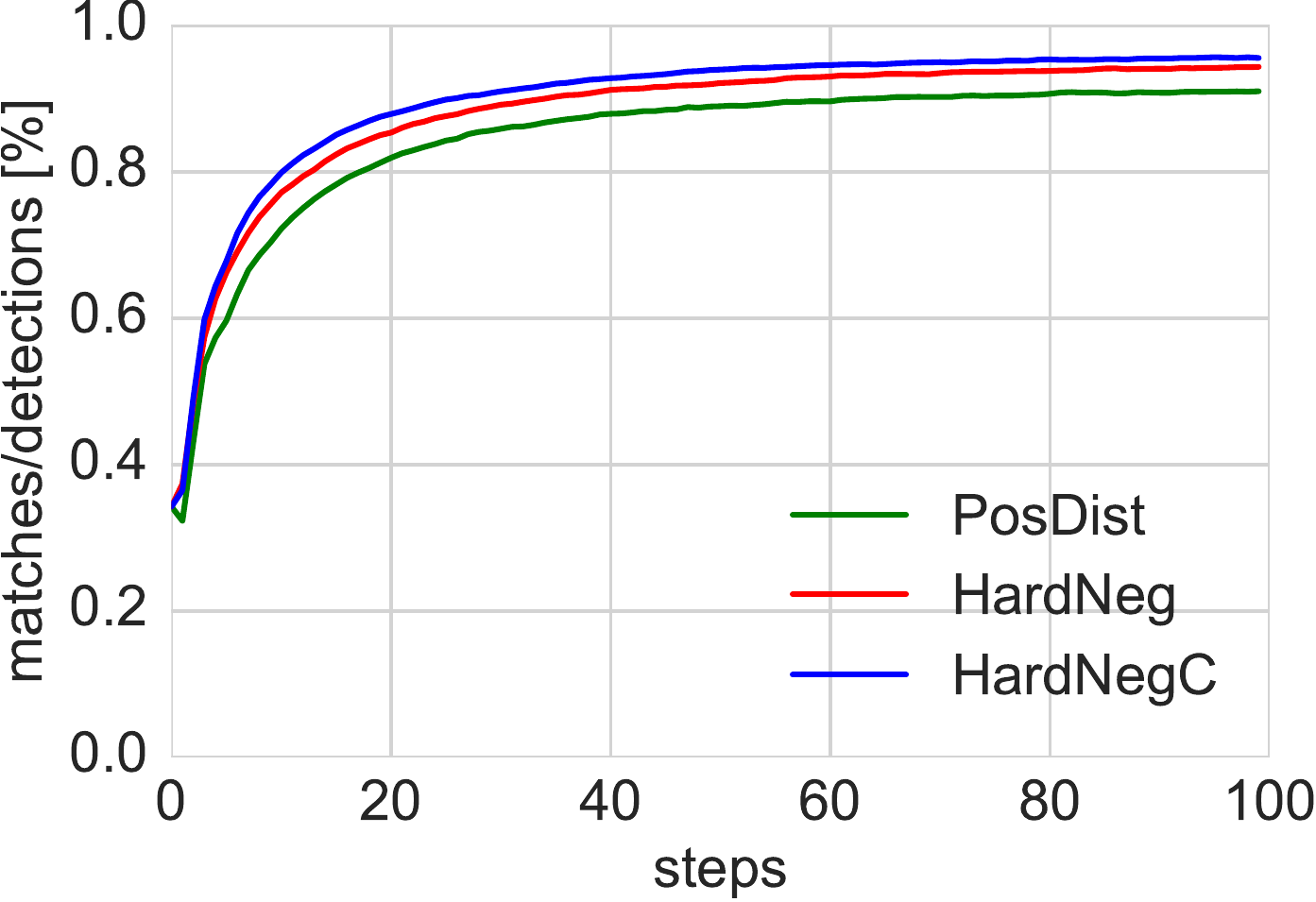}} SIFT \\
\end{minipage}
\hfill
\begin{minipage}[h]{0.24\linewidth}
\center{\includegraphics[width=1\linewidth]{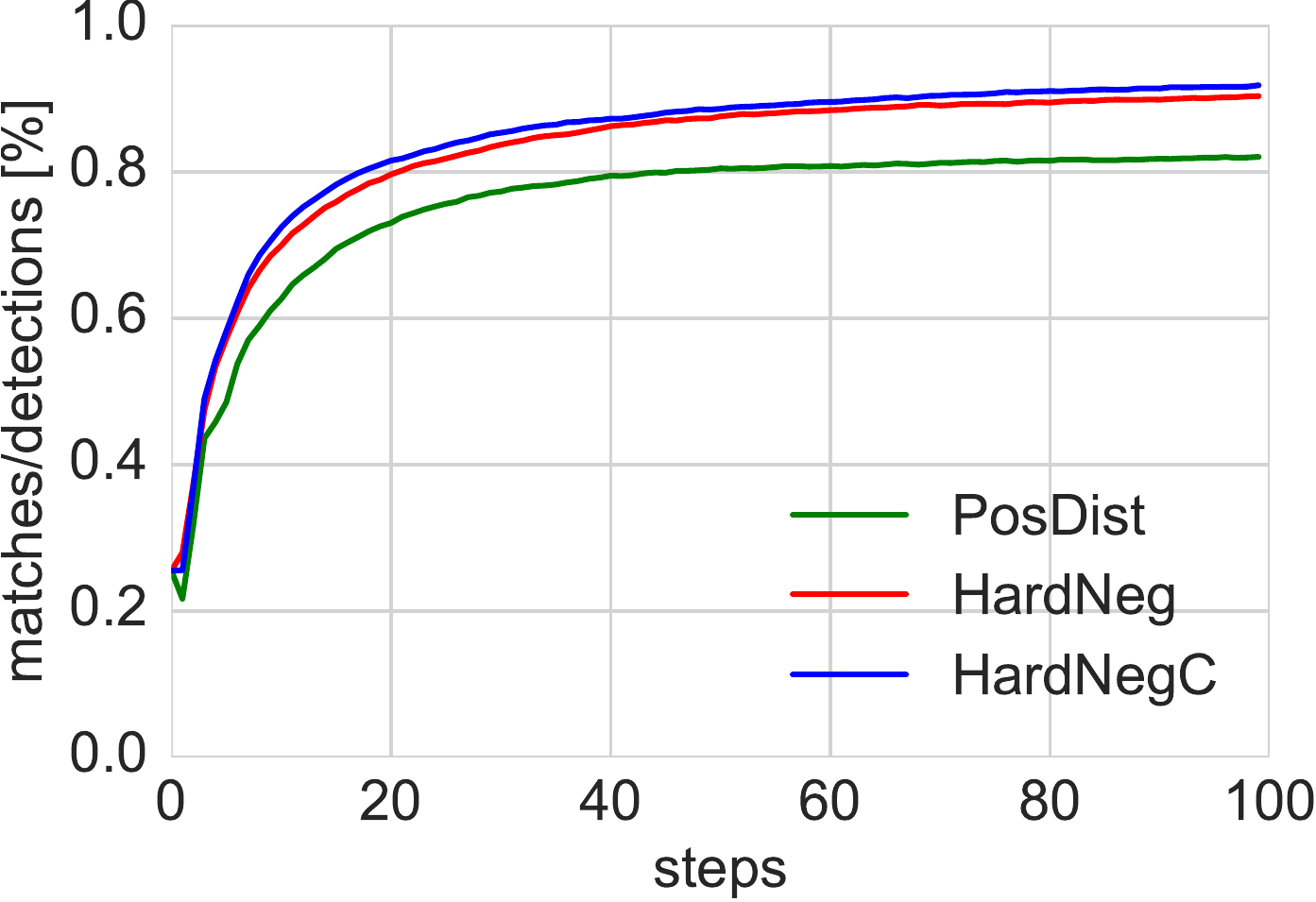}} Pixels \\
\end{minipage}

%% file: wxbstable.tex
\begin{table*}[htb]
\ra{1}
\centering
\caption{AffNet vs. Baumberg affine shape estimators on wide baseline stereo datasets, with Hessian and adaptive Hessian detectors, following the protocol~\cite{WXBS2015}. The number of matched image pairs and the average number of inliers. The \fbox{numbers} of image pairs in a dataset are boxed. Best results are in \textbf{bold}.
}
\setlength{\tabcolsep}{4pt}
\begin{tabular}{lrrrrrrrrrrrr}
\toprule
& \multicolumn{2}{c}{EF}
& \multicolumn{2}{c}{EVD}
& \multicolumn{2}{c}{OxAff}
& \multicolumn{2}{c}{SymB}
& \multicolumn{2}{c}{GDB}
& \multicolumn{2}{c}{LTLL}
\\
& \multicolumn{2}{c}{\cite{Zitnick2011}}
& \multicolumn{2}{c}{\cite{MODS2015}}
& \multicolumn{2}{c}{\cite{Mikolajczyk2005}}
& \multicolumn{2}{c}{\cite{Hauagge2012}}
& \multicolumn{2}{c}{\cite{Yang2007}}
& \multicolumn{2}{c}{\cite{Fernando2015}}
\\
\cmidrule(r){2-3}
\cmidrule(r){4-5}
\cmidrule(r){6-7}
\cmidrule(r){8-9}
\cmidrule(r){10-11}
\cmidrule(r){12-13}
Detector
 & \fbox{33}& inl.
 & \fbox{15}& inl. 
 & \fbox{40}& inl.
 & \fbox{46}& inl.
 & \fbox{22}& inl.
 & \fbox{172}& inl.\\
\cmidrule(r){1-13}
HesAff~\cite{Mikolajczyk2004} & \textbf{33} & 78 & \textbf{2} & 38 & \textbf{40} & 1008 & 34 & 153 & 17 & 199 &26 & 34  \\
HesAffNet & \textbf{33} & \textbf{112} & \textbf{2} & \textbf{48} & \textbf{40} & \textbf{1181}& \textbf{37} & \textbf{203} & \textbf{19} & \textbf{222}& \textbf{46} & \textbf{36}  \\
\cmidrule(r){1-13}
AdHesAff~\cite{WXBS2015} & \textbf{33} & 111 & 3 & 33 & \textbf{40} & 1330 & 35 & 190 & 19 & 286 & 28 & 35 \\
AdHesAffNet&   \textbf{33} & \textbf{165} & \textbf{4} & \textbf{42} & \textbf{40} & \textbf{1567} & \textbf{37} & \textbf{275} & \textbf{21} & \textbf{336} &  \textbf{48} & \textbf{39}\\
\bottomrule
\end{tabular}
\label{tab:wxbs-table}
\end{table*}